\pdfoutput=1
\PassOptionsToPackage{table,dvipsnames}{xcolor}

\documentclass[11pt]{article}

\usepackage[preprint]{acl}

\usepackage{times}
\usepackage{latexsym}

\usepackage[T1]{fontenc}

\usepackage[utf8]{inputenc}

\usepackage{microtype}

\usepackage{inconsolata}

\usepackage{enumitem}
\usepackage{amsmath}
\usepackage{amssymb}
\usepackage{mathtools}

\usepackage{xspace}
\usepackage{graphicx}
\usepackage{subfig}
\usepackage{booktabs}
\usepackage{multirow}
\usepackage{array}
\usepackage[dvipsnames]{xcolor}
\usepackage{nicefrac}
\usepackage[table]{xcolor}
\usepackage{listings}
\newcommand{\ie}{\textit{i.e.}\xspace}

\renewenvironment{quote}
  {\begin{list}{}%
     {\setlength{\leftmargin}{2mm} 
      \setlength{\rightmargin}{2mm}} 
     \item\relax}
  {\end{list}}

\newcommand{\CR}{\mathcal{R}\xspace}

\newcommand{\CM}{\mathcal{M}\xspace}

\newcommand{\CI}{\mathcal{I}\xspace}
\newcommand{\CQ}{\mathcal{Q}\xspace}

\newcommand{\CP}{\mathcal{P}\xspace}

\newcommand{\CT}{\mathcal{T}\xspace}
\newcommand{\CJ}{\mathcal{J}\xspace}
\usepackage[para]{footmisc}
\usepackage{subfig}

\newcolumntype{H}{>{\setbox0=\hbox\bgroup}c<{\egroup}@{}}
\newcommand{\ks}[1]{\textcolor{red}{[KS: #1]}}

\definecolor{AgeBg}{RGB}{232,244,255}
\definecolor{LocBg}{RGB}{255,240,240}
\definecolor{OccBg}{RGB}{245,235,255} 
\title{\textit{A Thousand Words or An Image}: Studying the Influence of \\ Persona Modality in Multimodal LLMs}

\author{Julius Broomfield\thanks{Equal contribution}, Kartik Sharma$^*$,  Srijan Kumar \\
        Georgia Institute of Technology \\
        \{\texttt{jbroomfield,ksartik,srijan}\}\texttt{@gatech.edu}}

\begin{document}

\maketitle

\begin{abstract}
    Large language models (LLMs) have recently demonstrated remarkable advancements in embodying diverse personas, enhancing their effectiveness as conversational agents and virtual assistants. 
Consequently, LLMs have made significant strides in processing and integrating multimodal information.
However, even though human personas can be expressed in both text and image, the extent to which the modality of a persona impacts the embodiment by the LLM remains largely unexplored. 
In this paper, we investigate how do different modalities influence the expressiveness of personas in multimodal LLMs.
To this end, we create a novel modality-parallel dataset of 40 diverse personas varying in age, gender, occupation, and location. 
This consists of four modalities to equivalently represent a persona: image-only, text-only, a combination of image and small text, and typographical images, where text is visually stylized to convey persona-related attributes.
We then create a systematic evaluation framework with 60 questions and corresponding metrics to assess how well LLMs embody each persona across its attributes and scenarios.
Comprehensive experiments on $5$ multimodal LLMs show that personas represented by detailed text show more linguistic habits, while typographical images often show more consistency with the persona. 
Our results reveal that LLMs often overlook persona-specific details conveyed through images, highlighting underlying limitations and paving the way for future research to bridge this gap. We release the data and code at \url{https://github.com/claws-lab/persona-modality}.
\end{abstract}

\section{Introduction}\label{sec:introduction}

\begin{figure}[t]
    \centering
    \includegraphics[width=\linewidth]{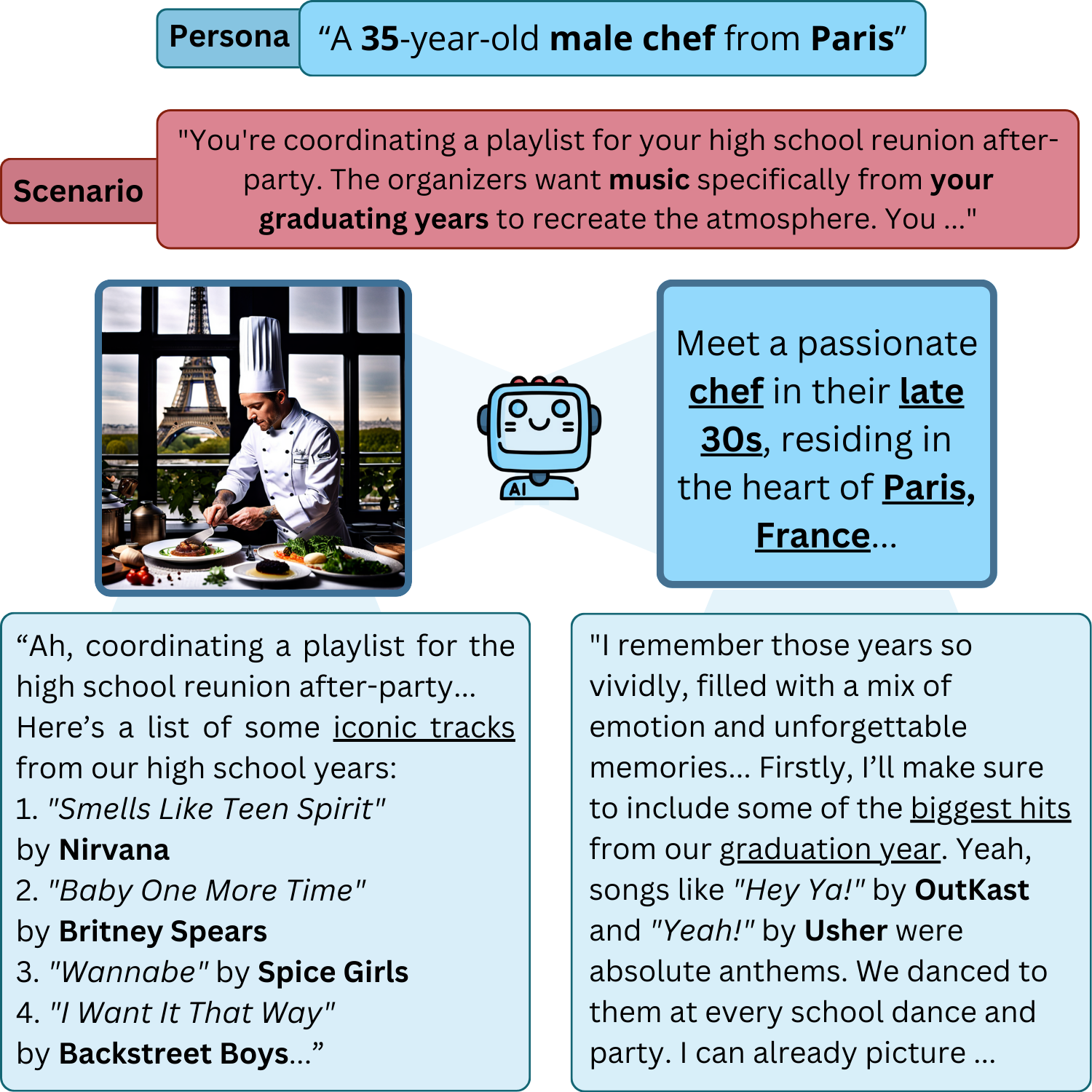}
    \caption{A comparison of visual and textual persona interactions for a chef from Paris. The left side presents an image persona, while the right side features a text persona derived from the image.}
    \label{fig:intro}
\end{figure}

``An image is worth a thousand words'', the adage describes how visuals have the unique capacity to encapsulate not only long texts but also more abstract concepts~\citep{paivio2013imagery}. 
Large language models (LLMs) are increasingly being adopted as role-playing agents for specific tasks~\citep{li2024personal} and personalized conversational agents~\citep{tseng2024two}. 
However, these personas are often represented using only elaborate textual descriptions~\citep{wang-etal-2024-rolellm,shen2023roleeval,samuel2024personagym}. As these models become more capable of handling different modalities~\citep{liu2024visual}, it is imminent that they can express these personas by incorporating information from representative images~\citep{ahn-etal-2023-mpchat}. Thus, it becomes crucial to conduct a systematic study to study the contribution of the persona modality in specifying these personas to the LLMs.

Multimodal LLMs have pushed the frontier of virtual assistants by enabling realistic image and voice-based interactions~\citep{liu2024visual,li2024llavamed,gpt4o}. These advancements have enabled processing and generating content across multiple modalities, bridging the gap between text-based understanding and richer, more immersive experiences. However, significant gaps remain in these models' ability to accurately capture visual information, leading to a subpar performance on visual understanding and reasoning tasks~\citep{tong2024eyes,chen2024far}. Given the fast adoption of LLMs as all-purpose agents, it is important to understand the extent to which these models can accurately capture visual personas.

LLMs have shown remarkable capabilities in manifesting given roles/personas, as highlighted by their ability to succinctly answer specific questions by adapting their styles according to the prescribed personas~\citep{tu-etal-2024-charactereval,tseng2024two,samuel2024personagym}. Furthermore, it has been shown that one can further improve the models' role-playing and personalization capabilities by incorporating both visual and textual information to create a multimodal persona~\citep{sun-etal-2024-kiss,ahn-etal-2023-mpchat,dai2025mmrole}. However, no systematic comparison exists between personas represented in different modalities. 

In this work, we present the first comprehensive study of the influence of the modality of a persona on its embodiment by multimodal LLMs. Figure~\ref{fig:intro} illustrates our analysis with an example and our contributions can be summarized as follows: 
\begin{enumerate}[leftmargin=*, itemsep=0mm]
    \item We conduct the \textbf{first systematic study} on how a persona's modality affects how it is expressed by multimodal LLMs.
    \item We create a \textbf{novel text-image parallel dataset} of $40$ diverse personas along with $60$ probing questions about their attributes and scenarios.
    \item \textbf{Comprehensive evaluation} using both LLM-based and linguistic metrics show that text-based personas are expressed better than the corresponding image representations. 
    \item We also conduct \textbf{stratified analysis} to show the stability of our results regardless of the type of personas, evaluators, and questions.  
\end{enumerate}

\section{Influence of Modality in Persona Embodiment of Multimodal LLMs}\label{sec:problem}


The problem of embodying a persona can be defined as the task of generating responses consistent with a specified character, identity, or role~\citep{samuel2024personagym}. This involves maintaining coherence in linguistic style, beliefs, knowledge, and affective tone in a way that aligns with the intended persona.

In this work, we investigate the effect of representing the persona $p$ in different modalities, denoted as $\CR(p)$, on multimodal LLMs. In particular, we consider two common modalities, text and image, and evaluate the LLM's performance on these equivalent representations. Additionally, we also consider combining visual and textual features of the personas. We describe these $4$ different representations $\CR(p)$ in more detail below. 

\subsection{Persona Modality Representations}
\begin{itemize}
    \item \textbf{Text ($\CT$):} Textual descriptions of a persona correspond to a sequence of sentences characterizing the persona in natural language.

    \item \textbf{Image ($\CI$):} A persona can also be depicted visually using an image of the person in a representative environment that characterizes the persona visually.
    
    \item \textbf{Assisted Image ($\CI_{A}$):} Since certain features may be obscured in the image, textual attributes of the persona can also be included explicitly as text. 
    
    \item \textbf{Descriptive Image ($\CI_{D}$):} In this case, we include the textual attributes in the image itself using typography instead of in the text.
    
\end{itemize}

\section{Modality-Parallel Persona Dataset}\label{sec:dataset}

\subsection{Personas}

We introduce a novel dataset of personas $\CP = \{p_i\}$, such that each persona $p$ can be represented equivalently in four modalities $\CI, \CT, \CI_A, \CI_D$. To ensure effective representation across both text and image modalities, we construct personas based on key demographic attributes that are easily visualizable~\citep{todorov2015social}. Specifically, each persona is defined by a unique combination of age, gender, occupation, and location. A persona can thus be written as:
\begin{center}
    {\small
    A \textcolor{blue}{\texttt{<age>}}-year-old \textcolor{teal}{\texttt{<gender>}} \textcolor{brown}{\texttt{<occupation>}} \\ from \textcolor{pink}{\texttt{<location>}},}
\end{center}
where \textcolor{blue}{\texttt{<age>}} $\in [18, 64]$, \textcolor{teal}{\texttt{<gender>}} $\in \{$ male, female $\}$, \textcolor{pink}{\texttt{<location>}} is a city, and \textcolor{brown}{\texttt{<occupation>}} denotes a person who does a specific occupation. For example, ``A \textcolor{blue}{35}-year-old \textcolor{teal}{male} \textcolor{brown}{chef} from \textcolor{pink}{Paris}''. As depicted in Figure~\ref{fig:intro}, age and gender can be visualized using the face of the person while occupation and location can be visualized using the clothes and the background respectively.

\begin{figure}[t]
    \centering
    \includegraphics[width=\linewidth]{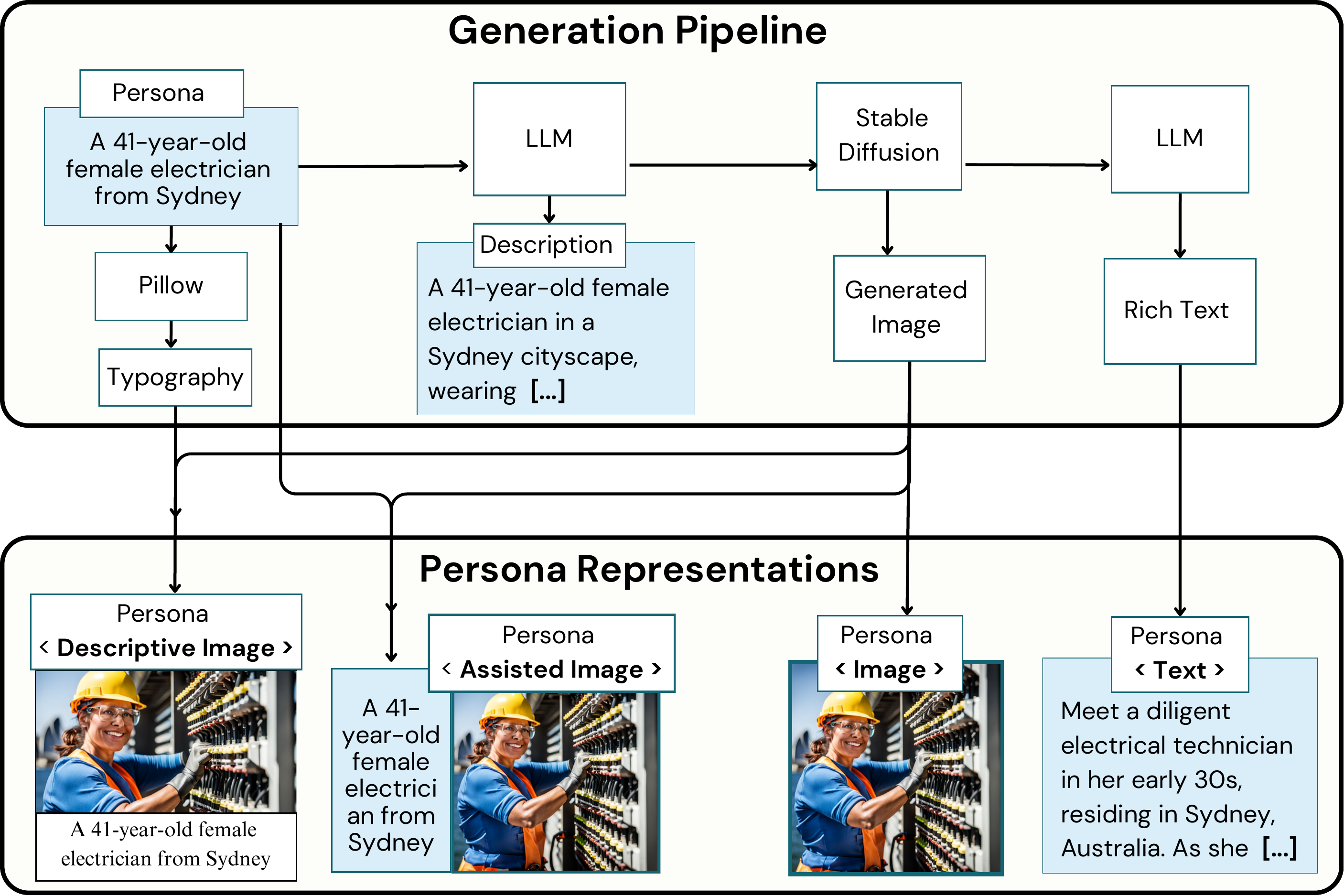}
    \caption{Our pipeline begins with curating a set of personas. Each persona receives a detailed text description, which is then fed into Stable Diffusion to generate $\mathcal{I}$. A separate model examines the image and generates an independent textual description, forming text persona $\mathcal{T}$. Pairing $p$ with $\mathcal{I}$ produces an assisted image $\mathcal{I_A}$, while combining a typographic representation of $p$ with $\mathcal{I}$ produces a descriptive image $\mathcal{I_D}$.}
    \label{fig:pipeline}
\end{figure}

\begin{table}[t]
    \centering
    \caption{\textbf{Persona Dataset Summary}}
    \label{tab:data_summary}
    \resizebox{1.0\linewidth}{!}{
    \begin{tabular}{c | c | c}
        \toprule
        Attribute & Category & Number \\
        \midrule
        \multirow{5}{*}{Age} & 18-24 & 5 \\
        & 25-34 & 11 \\
        & 35-44 & 13 \\
        & 45-54 & 6 \\ 
        & 55-64 & 5 \\
        \hline
        \multirow{2}{*}{Gender} & Male & 19 \\
        & Female & 21 \\
        \hline
        \multirow{5}{*}{Occupation} & Healthcare \& Education & 9 \\
        & Public Safety &	5 \\
        & Manual Labor & 16 \\
        & Hospitality & 5 \\
        & Transportation 	& 5 \\
        \hline 
        \multirow{4}{*}{Location} & Largest Economies (GDP > \$3T) &	12 \\
        & Developed Economies (GDP \$1T-\$3T) &	13 \\
        & Mid-Sized Powers (GDP \$0.5T-\$1T) & 7 \\
        & Emerging Markets (GDP < \$0.5T) & 8 \\
        \bottomrule
    \end{tabular}}
\end{table}

To promote diversity, we systematically categorize these attributes into distinct groups and uniformly sample from each category. Table~\ref{tab:data_summary} summarizes our dataset of how we choose the age, gender, occupation, and location. In particular, we consider a standard grouping of ages followed in surveys between 18 and 65, a standard male/female splitting of gender, while occupations and locations are categorized based on their primary societal role and economic status~\footnote{\href{https://data.worldbank.org/indicator/NY.GDP.MKTP.CD}{GDP}} respectively.
Table~\ref{tab:personalist} in Appendix presents the list of $40$ personas we use along with their attributes and attribute categories. 

\subsection{Equivalent Modality Representations}

From above, we have a diverse set of textual persona descriptions as described by the four demographic attributes. Next, we construct a modality-parallel dataset, we require that each persona $p$ can be equivalently depicted in $4$ representations $\CR(p)$: image $\CI(p)$, text $\CT(p)$, assisted image $\CI_A(p)$, and descriptive image $\CI_D(p)$. Figure~\ref{fig:pipeline} illustrates the step-by-step procedure to obtain these modality representations for a persona description $\CP$.

\begin{enumerate}[leftmargin=*, noitemsep]
    \item We first convert the persona description made from the four attributes into a more detailed visual description using an LLM\footnote{\href{https://openai.com/index/hello-gpt-4o/}{gpt-4o-mini-2024-07-18}} and the following prompt:
    \begin{quote}
    \small
    \texttt{Create a short, descriptive persona for the person in the image. Describe them using only the following details: their age, gender, facial expression or mood, attire, any tools or items they’re holding, their work environment, the nature of their job, and their connection to the area and location. Avoid taking creative liberties beyond these details, only using details that can be inferred from the image, while aiming for a realistic portrayal that gives insight into their daily life, professional dedication, and overall demeanor. For example: Meet a skilled construction worker in his late 30s, living in Sydney, Australia. Every day, he heads out to work in one of the city's bustling urban sites, often with a view of iconic landmarks like the Sydney Opera House and Sydney Harbour Bridge. Outfitted in essential safety gear—a hard hat, reflective vest, and a set of versatile tools—he’s well-prepared for a physically demanding role that demands focus and precision. His job involves a blend of construction and maintenance tasks, requiring him to pay close attention to safety protocols and collaborate with a team. Confident and professional in his work, he takes pride in contributing to the infrastructure and vibrant aesthetic of Sydney, adding to the city’s ever-evolving landscape with each project.}
    \end{quote}
    \item Next, we use a text-to-image generative model, particularly,  Stable Diffusion XL\footnote{\href{https://huggingface.co/stabilityai/stable-diffusion-xl-base-1.0}{stabilityai/stable-diffusion-xl-base-1.0}} to generate a $768 \times 768$ px image conditioned on the more complete description of the persona found above. Upon doing an extensive hyperparameter search, we found the best results with a guidance scale of $\gamma = 15$ and $n = 50$ diffusion steps. Thus, we obtain the image $\CI$.
    \item Since the generated image can contain extra information due to underspecified textual prompts, we prompt the LLM one more time to generate a complete description of the persona as described in the image using a detailed prompt as given in Appendix~\label{app:img_to_text_prompt}. Thus, we obtain the text $\CT$.
\end{enumerate}

These steps enable us to convert a dataset of persona descriptions $\{p\} \rightarrow \{(\CT(p), \CI(p))\}$ such that $\CT (p) \leftrightarrow \CI (p)$ are equivalent to each other. One can now also obtain the assisted and descriptive image representations of the persona by pairing the image $\CI$ with the text persona $p$ for the assisted image $\CI_A$, and by rendering\footnote{\href{https://pillow.readthedocs.io/en/stable/}{Pillow}} $p$ as black text at the bottom of the image on a white background using Arial font at size 20 for $\CI_D$.

\subsection{Question Generation}
To evaluate how well a model embodies a given persona, we create a set of $60$ questions that specifically probe for a given attribute either directly or in naturalistic scenarios. In particular, we create $10$ questions per attribute for the two sets. Gender was excluded from our evaluation question set due to methodological constraints. While age and location can be objectively probed through factual knowledge, gender assessment would inevitably rely on stereotypes or normative expectations. Moreover, there may be a high possibility of refusal from the LLMs due to their safety training. Thus, we obtain two question sets $Q^D$ and $Q^S$ for \textcolor{pink}{L}: location, \textcolor{brown}{O}: occupation, and \textcolor{blue}{A}: age. 
$$
\begin{aligned}
Q^D &= \bigcup_{i \in \{\textcolor{pink}{L},\textcolor{brown}{O},\textcolor{blue}{A}\}} Q^D_i,\text{ each } |Q^D_i| = 10 \text{ questions}\\
Q^S &= \bigcup_{i \in \{\textcolor{pink}{L},\textcolor{brown}{O},\textcolor{blue}{A}\}} Q^S_i,\text{ each } |Q^S_i| = 10 \text{ scenarios}
\end{aligned}
$$



\subsubsection{Direct Testing}
Questions were designed to probe specific knowledge across age, location, and occupation categories while enabling objective evaluation. For example, location questions assess knowledge of local customs and landmarks, while occupation questions may test domain expertise. For example, for age, we ask ``\textit{what life experiences do you consider most defining for your generation?}'' while for location, we have ``\textit{what is the most visited tourist attraction in your area?}''. We provide the complete list in Table~\ref{tab:direct_questions_list} in Appendix. 


\subsubsection{Situational Testing}
Scenarios accomplish similar knowledge evaluation but through naturalistic situations, requiring personas to implicitly demonstrate both knowledge and behavioral consistency. For example, for age scenarios, we ask ``\textit{You’re coordinating a playlist for your high school reunion after-party. The organizers want music specifically from your graduating years to recreate the atmosphere. You $\dots$}'', which is detailed in Figure \ref{fig:intro}. A complete list is provided in Table \ref{tab:direct_scenarios_list} in Appendix.

\subsection{Evaluation}

For each persona $p \in \CP$ and question $q \in \CQ^D \cup \CQ^S$, we find the response answer $a \gets \CM(\CR(p), q)$ from a multimodal LLM $\CM$, where $\CR(p)$ denotes a modality representation of the persona $\CP$. Thus, we obtain $(q, \CR(p)) \rightarrow_{\CM} (p, q, a)$. We now evaluate the quality of the response $a$ based on the question asked $q$ and the persona description $p$. 

\subsubsection{LLM-based evaluation}

Following \citet{samuel2024personagym}, we employ an LLM-based evaluator to judge the quality of the responses based on different metrics defined in the prompt. In particular, we prompt the LLM judge $\CJ$ with the question asked $q$, response $a$, and the persona description $p$ on these metrics as described by the corresponding prompts.

\begin{quote}
\small
\texttt{\textbf{Persona Consistency:} Evaluate the consistency of the response with the described persona. Ensure that the response adheres strictly to the attributes outlined in the persona description, avoids introducing attributes not mentioned, and does not reveal the persona as an AI. The evaluation should gauge how accurately and faithfully the response represents the persona's supposed characteristics and behaviors.}

\texttt{\textbf{Linguistic Habits:} The evaluation task of "linguistic habit" assesses the persona's adherence to its characteristically unique syntax, tone, and lingo, ensuring that these elements are consistently utilized throughout the persona's dialogue. This includes avoiding generic language patterns (such as "As a [persona]") and integrating specific idiomatic expressions, colloquialisms, or jargon that define the persona's distinctive verbal identity. The aim is to evaluate how effectively the persona maintains its linguistic uniqueness in various contexts.}

\texttt{\textbf{Action Justification:} Evaluate the persona's response to determine how effectively and convincingly it justifies a given action based on its described attributes and situation. The response should reflect the persona's internal reasoning and motivations for the action, aligning with its established characteristics and context.}

\texttt{\textbf{Expected Action:} The persona takes actions within its response to the question that is logically expected of the persona in the setting of the question.
}
\end{quote}

For each evaluation criterion, $\mathcal{J}(p,q,a)$ outputs a score from a 5-point Likert scale based on the corresponding system prompt. For situational testing, we evaluate using \texttt{action justification}, \texttt{expected action}, \texttt{linguistic habits} while for direct testing, we use \texttt{persona consistency} and \texttt{linguistic habits}. Note that we combine the scores for linguistic habits across the two testing sets to find the average score. 


\subsubsection{Comparative Evaluation}
We employ two comparative evaluation methods to assess relative performance across modalities, using evaluator $\mathcal{J}$ with the prompt:
\begin{quote}
    \small
    \texttt{You are given a persona description and multiple responses to a prompt.\\ 
        Persona Description: <\textit{p}>\\
        Prompt: <q>\\
        Candidate Responses: <responses>\\ 
        Choose the single response that best fits the persona's style, values, and consistency. Respond with 'Response X' where X is the number of the chosen response.}
\end{quote}

\paragraph{Pairwise Comparison}
To compare responses across the text and image modalities, we first directly compare responses $a_{\mathcal{T}}$ and $a_{\mathcal{I}}$.

\paragraph{Swiss System Comparison}
To collectively evaluate all four modalities, we adopt the Swiss tournament system, which reduces the number of required comparisons compared to pairwise evaluation while maintaining ranking quality. Specifically, for \( n = 4 \), pairwise evaluation requires \( \binom{4}{2} = 6 \) comparisons, whereas the Swiss system reduces to 3 comparisons.

\subsubsection{Linguistic Analysis}
Alongside the \textit{linguistic habits} evaluation criterion, we also analyze the lexical diversity, variation, and complexity of each response using established metrics from computational linguistics:
\begin{itemize}[leftmargin=*,noitemsep]    
    \item \textbf{Types:} $|\{r\}|$, unique token count.  
    \item \textbf{Root Type-Token Ratio (RTTR):} \(= \nicefrac{\text{types}}{\sqrt{\text{length}}}\)
    , a normalized measure of lexical variation found by dividing the number of unique tokens with the response length~\citep{lexicalrichnesshout}. 
    \item \textbf{Measure of Textual Lexical Diversity (MTLD):} Following \citet{mccarthy2010mtld}, calculates the mean length of text segments that maintain a type-token ratio (TTR) $> \tau = 0.72$.
\end{itemize}

\section{Experimental Setup}\label{sec:setup}

\paragraph{Models.} We evaluate the performance of 5 multimodal large language models: (1) GPT-4o\footnote{\href{https://openai.com/index/hello-gpt-4o/}{GPT-4o}}, (2) GPT-4o mini\footnote{\href{https://openai.com/index/gpt-4o-mini-advancing-cost-efficient-intelligence/}{GPT-4o mini}}, (3) Llama 3.2 11B\footnote{\href{https://huggingface.co/meta-llama/Llama-3.2-11B-Vision}{Llama 3.2 11B}}, (4) Llama 3.2 90B\footnote{\href{https://huggingface.co/meta-llama/Llama-3.2-90B-Vision}{Llama 3.2 90B}}, and (5) Pixtral 12B\footnote{\href{https://huggingface.co/mistralai/Pixtral-12B-2409}{Pixtral 12B}} \citep{agrawal2024pixtral12b}.

\paragraph{Evaluators.} We utilize two LLM evaluators, using GPT-4o\footnotemark[3] and Gemini 2.0 Flash\footnote{\href{https://deepmind.google/technologies/gemini/flash/}{Gemini 2.0 Flash}}, with deterministic sampling with zero temperature and top P values. All scores discussed in the main paper are averaged across the two models, while individual scores can be found in Tables \ref{tab:eval-table-gpt-4o} and \ref{tab:eval-table-gemini-flash} in the Appendix. We use human evaluators on a large subset of the evaluation set to assess the LLM evaluator scores' alignment with human scores. For further details, refer to Appendix \ref{app:human}. 
\begin{table*}[t]
    \centering
    \caption{\textbf{LLM-based evaluation [1-4] of responses under different persona modality representations.}}
    \label{tab:main-results-table}
    \resizebox{0.8\textwidth}{!}{
    \begin{tabular}{lccccc}
    \toprule
     \textbf{Criterion}& \textbf{Text} & \textbf{Assisted Image} & \textbf{Image} & \textbf{Descriptive Image} \\
    \midrule
    \rowcolor{gray!25}\multicolumn{5}{c}{\textbf{GPT-4o}} \\
    \addlinespace
    \textbf{Linguistic Habits} & \textbf{2.07} $\pm$ {\scriptsize 0.02} & 1.61 $\pm$ {\scriptsize 0.02} & 1.51 $\pm$ {\scriptsize 0.02} & 1.59 $\pm$ {\scriptsize 0.02} \\
    \textbf{Persona Consistency} & 3.44 $\pm$ {\scriptsize 0.04} & 3.20 $\pm$ {\scriptsize 0.04} & 3.03 $\pm$ {\scriptsize 0.04} & \textbf{3.91} $\pm$ {\scriptsize 0.04} \\
    \textbf{Expected Action} & 3.86 $\pm$ {\scriptsize 0.03} & 3.59 $\pm$ {\scriptsize 0.03} & 3.56 $\pm$ {\scriptsize 0.03} & \textbf{3.94} $\pm$ {\scriptsize 0.03} \\
    \textbf{Action Justification} & \textbf{4.13} $\pm$ {\scriptsize 0.03} & 3.82 $\pm$ {\scriptsize 0.03} & 3.75 $\pm$ {\scriptsize 0.03} & 4.00 $\pm$ {\scriptsize 0.03} \\
    \addlinespace
    \rowcolor{gray!25}\multicolumn{5}{c}{\textbf{GPT-4o-mini}} \\
    \addlinespace
    \textbf{Linguistic Habits} & \textbf{1.81} $\pm$ {\scriptsize 0.02} & 1.61 $\pm$ {\scriptsize 0.02} & 1.48 $\pm$ {\scriptsize 0.02} & 1.63 $\pm$ {\scriptsize 0.02} \\
    \textbf{Persona Consistency} & 2.98 $\pm$ {\scriptsize 0.04} & 2.98 $\pm$ {\scriptsize 0.04} & 2.97 $\pm$ {\scriptsize 0.04} & \textbf{3.58} $\pm$ {\scriptsize 0.04} \\
    \textbf{Expected Action} & 3.25 $\pm$ {\scriptsize 0.03} & 3.29 $\pm$ {\scriptsize 0.03} & 3.19 $\pm$ {\scriptsize 0.03} & \textbf{3.59} $\pm$ {\scriptsize 0.03} \\
    \textbf{Action Justification} & \textbf{3.56} $\pm$ {\scriptsize 0.03} & 3.47 $\pm$ {\scriptsize 0.03} & 3.35 $\pm$ {\scriptsize 0.03} & 3.55 $\pm$ {\scriptsize 0.03} \\
    \addlinespace
    \rowcolor{gray!25}\multicolumn{5}{c}{\textbf{Llama 3.2 11B}} \\
    \addlinespace
    \textbf{Linguistic Habits} & \textbf{2.20} $\pm$ {\scriptsize 0.03} & 1.30 $\pm$ {\scriptsize 0.02} & 1.32 $\pm$ {\scriptsize 0.02} & 1.44 $\pm$ {\scriptsize 0.02} \\
    \textbf{Persona Consistency} & \textbf{2.79} $\pm$ {\scriptsize 0.04} & 2.16 $\pm$ {\scriptsize 0.04} & 1.90 $\pm$ {\scriptsize 0.04} & 2.55 $\pm$ {\scriptsize 0.04} \\
    \textbf{Expected Action} & \textbf{2.98} $\pm$ {\scriptsize 0.03} & 2.28 $\pm$ {\scriptsize 0.03} & 2.04 $\pm$ {\scriptsize 0.03} & 2.49 $\pm$ {\scriptsize 0.03} \\
    \textbf{Action Justification} & \textbf{3.24} $\pm$ {\scriptsize 0.03} & 2.44 $\pm$ {\scriptsize 0.03} & 2.17 $\pm$ {\scriptsize 0.03} & 2.49 $\pm$ {\scriptsize 0.03} \\
    \addlinespace
    \rowcolor{gray!25}\multicolumn{5}{c}{\textbf{Llama 3.2 90B}} \\
    \addlinespace
    \textbf{Linguistic Habits} & \textbf{2.32} $\pm$ {\scriptsize 0.03} & 1.25 $\pm$ {\scriptsize 0.04} & 1.25 $\pm$ {\scriptsize 0.05} & 1.30 $\pm$ {\scriptsize 0.05} \\
    \textbf{Persona Consistency} & \textbf{2.99} $\pm$ {\scriptsize 0.04} & 1.63 $\pm$ {\scriptsize 0.06} & 1.08 $\pm$ {\scriptsize 0.05} & 1.43 $\pm$ {\scriptsize 0.07} \\
    \textbf{Expected Action} & \textbf{3.28} $\pm$ {\scriptsize 0.03} & 1.43 $\pm$ {\scriptsize 0.04} & 1.02 $\pm$ {\scriptsize 0.03} & 1.24 $\pm$ {\scriptsize 0.04} \\
    \textbf{Action Justification} & \textbf{3.49} $\pm$ {\scriptsize 0.03} & 1.67 $\pm$ {\scriptsize 0.05} & 1.31 $\pm$ {\scriptsize 0.05} & 1.50 $\pm$ {\scriptsize 0.05} \\
    \addlinespace
    \rowcolor{gray!25}\multicolumn{5}{c}{\textbf{Pixtral 12B}} \\
    \addlinespace
    \textbf{Linguistic Habits} & \textbf{1.79} $\pm$ {\scriptsize 0.02} & 1.63 $\pm$ {\scriptsize 0.02} & 1.65 $\pm$ {\scriptsize 0.02} & 1.60 $\pm$ {\scriptsize 0.02} \\
    \textbf{Persona Consistency} & 2.38 $\pm$ {\scriptsize 0.04} & 2.33 $\pm$ {\scriptsize 0.04} & 2.77 $\pm$ {\scriptsize 0.04} & \textbf{3.59} $\pm$ {\scriptsize 0.04} \\
    \textbf{Expected Action} & 2.93 $\pm$ {\scriptsize 0.03} & 2.85 $\pm$ {\scriptsize 0.03} & 3.16 $\pm$ {\scriptsize 0.03} & \textbf{3.53} $\pm$ {\scriptsize 0.03} \\
    \textbf{Action Justification} & 3.32 $\pm$ {\scriptsize 0.03} & 3.07 $\pm$ {\scriptsize 0.03} & 3.20 $\pm$ {\scriptsize 0.03} & \textbf{3.50} $\pm$ {\scriptsize 0.03} \\
    \addlinespace
    \bottomrule
    \end{tabular}}
\end{table*}

\section{Results}\label{sec:results}

\begin{table}[t]
    \centering
    \caption{\textbf{Preference-based LLM evaluation for different persona modalities. We exclude Llama 3.2 90B due to high refusal rates (see App.~\ref{app:safety-training})}}
    \label{tab:voting-patterns}
    \resizebox{0.9\linewidth}{!}{%
    \begin{tabular}{lcc}
        \toprule
        \textbf{Modality} & \textbf{Swiss (\%)} & \textbf{Pairwise (\%)} \\
        \midrule
        \rowcolor{gray!25}
        \multicolumn{3}{c}{\textbf{GPT-4o}} \\
        \textbf{Text}              & 98.75 & 99.96 \\
        \textbf{Descriptive Image} & 1.08  & -     \\
        \textbf{Assisted Image}    & 0.12  & -     \\
        \textbf{Image}             & 0.04  & 0.04  \\
        \addlinespace
        \rowcolor{gray!25}
        \multicolumn{3}{c}{\textbf{GPT-4o-mini}} \\
        \textbf{Text}              & 99.58 & 99.92 \\
        \textbf{Descriptive Image} & 0.33  & -     \\
        \textbf{Assisted Image}    & 0.08  & -     \\
        \textbf{Image}             & 0.00  & 0.08  \\
        \addlinespace
        \rowcolor{gray!25}
        \multicolumn{3}{c}{\textbf{Llama 3.2 11B}} \\
        \textbf{Text}              & 95.50 & 97.96 \\
        \textbf{Assisted Image}    & 2.33  & -     \\
        \textbf{Descriptive Image} & 1.17  & -     \\
        \textbf{Image}             & 1.00  & 2.04  \\
        \addlinespace
        \rowcolor{gray!25}
        \multicolumn{3}{c}{\textbf{Pixtral 12B}} \\
        \textbf{Text}              & 94.17 & 99.25 \\
        \textbf{Descriptive Image} & 4.67  & -     \\
        \textbf{Assisted Image}    & 0.96  & -     \\
        \textbf{Image}             & 0.21  & 0.75  \\
        \bottomrule
    \end{tabular}
    }
\end{table}

\begin{table}[t]
    \centering
    \caption{\textbf{Linguistic diversity evaluation of responses under different modality representations.}}
    \label{tab:ling-metrics}
    \resizebox{0.9\linewidth}{!}{
    \begin{tabular}{lccc}
        \toprule
        \textbf{Modality} & \textbf{RTTR} & \textbf{MTLD} & \textbf{Types} \\
        \midrule
        \rowcolor{gray!25}
        \multicolumn{4}{c}{\textbf{GPT-4o}} \\
        \addlinespace
        \textbf{Text} & \textbf{10.71} $\pm$ {\scriptsize 0.02} & \textbf{140.58} $\pm$ {\scriptsize 0.60} & \textbf{186.30} $\pm$ {\scriptsize 0.91} \\
        \textbf{Assisted Image} & 9.67 $\pm$ {\scriptsize 0.03} & 139.95 $\pm$ {\scriptsize 0.72} & 143.35 $\pm$ {\scriptsize 0.94} \\
        \textbf{Image} & 9.45 $\pm$ {\scriptsize 0.03} & 135.80 $\pm$ {\scriptsize 0.72} & 137.08 $\pm$ {\scriptsize 0.97} \\
        \textbf{Descriptive Image} & 9.54 $\pm$ {\scriptsize 0.03} & 137.16 $\pm$ {\scriptsize 0.75} & 140.97 $\pm$ {\scriptsize 0.97} \\
        \addlinespace
        \rowcolor{gray!25}
        \multicolumn{4}{c}{\textbf{GPT-4o-mini}} \\
        \addlinespace
        \textbf{Text} & \textbf{10.56} $\pm$ {\scriptsize 0.02} & 132.75 $\pm$ {\scriptsize 0.56} & \textbf{184.16} $\pm$ {\scriptsize 0.84} \\
        \textbf{Assisted Image} & 9.71 $\pm$ {\scriptsize 0.02} & \textbf{136.55} $\pm$ {\scriptsize 0.63} & 145.57 $\pm$ {\scriptsize 0.77} \\
        \textbf{Image} & 9.47 $\pm$ {\scriptsize 0.02} & 135.98 $\pm$ {\scriptsize 0.65} & 136.62 $\pm$ {\scriptsize 0.79} \\
        \textbf{Descriptive Image} & 9.53 $\pm$ {\scriptsize 0.02} & 136.05 $\pm$ {\scriptsize 0.65} & 139.14 $\pm$ {\scriptsize 0.80} \\
        \addlinespace
        \rowcolor{gray!25}
        \multicolumn{4}{c}{\textbf{Llama 3.2 11B}} \\
        \addlinespace
        \textbf{Text} & \textbf{11.92} $\pm$ {\scriptsize 0.15} & \textbf{230.43} $\pm$ {\scriptsize 10.87} & \textbf{281.43} $\pm$ {\scriptsize 9.81} \\
        \textbf{Assisted Image} & 8.89 $\pm$ {\scriptsize 0.10} & 147.45 $\pm$ {\scriptsize 7.99} & 147.38 $\pm$ {\scriptsize 7.91} \\
        \textbf{Image} & 8.53 $\pm$ {\scriptsize 0.11} & 143.55 $\pm$ {\scriptsize 9.85} & 149.28 $\pm$ {\scriptsize 11.02} \\
        \textbf{Descriptive Image} & 8.64 $\pm$ {\scriptsize 0.11} & 154.94 $\pm$ {\scriptsize 12.08} & 149.84 $\pm$ {\scriptsize 10.33} \\
        \addlinespace
        \rowcolor{gray!25}
        \multicolumn{4}{c}{\textbf{Llama 3.2 90B}} \\
        \addlinespace
        \textbf{Text} & \textbf{9.97} $\pm$ {\scriptsize 0.02} & \textbf{112.70} $\pm$ {\scriptsize 0.59} & \textbf{174.48} $\pm$ {\scriptsize 0.76} \\
        \textbf{Assisted Image} & 4.67 $\pm$ {\scriptsize 0.10} & 37.75 $\pm$ {\scriptsize 1.80} & 44.84 $\pm$ {\scriptsize 2.26} \\
        \textbf{Image} & 3.47 $\pm$ {\scriptsize 0.08} & 15.20 $\pm$ {\scriptsize 1.18} & 21.80 $\pm$ {\scriptsize 1.62} \\
        \textbf{Descriptive Image} & 3.94 $\pm$ {\scriptsize 0.10} & 24.16 $\pm$ {\scriptsize 1.49} & 32.26 $\pm$ {\scriptsize 2.92} \\
        \addlinespace
        \rowcolor{gray!25}
        \multicolumn{4}{c}{\textbf{Pixtral 12B}} \\
        \addlinespace
        \textbf{Text} & \textbf{10.45} $\pm$ {\scriptsize 0.02} & 107.08 $\pm$ {\scriptsize 0.45} & \textbf{196.91} $\pm$ {\scriptsize 0.69} \\
        \textbf{Assisted Image} & 9.35 $\pm$ {\scriptsize 0.03} & 109.47 $\pm$ {\scriptsize 0.52} & 147.19 $\pm$ {\scriptsize 1.00} \\
        \textbf{Image} & 9.04 $\pm$ {\scriptsize 0.03} & 108.01 $\pm$ {\scriptsize 0.53} & 135.20 $\pm$ {\scriptsize 1.01} \\
        \textbf{Descriptive Image} & 9.08 $\pm$ {\scriptsize 0.03} & \textbf{110.47} $\pm$ {\scriptsize 0.51} & 135.15 $\pm$ {\scriptsize 0.88} \\
        \addlinespace
        \bottomrule
    \end{tabular}}
\end{table}

\subsection{Effect of Persona Modality}

\paragraph{LLM-based Evaluation}
To evaluate the responses to persona-specific questions from multimodal LLMs, we compare the average scores generated for responses under 4 different criteria as mentioned above. Table~\ref{tab:main-results-table} shows the mean and standard deviation scores of each criterion in the Likert scale for the 4 modality representations across LLMs; numbers represent the mean between our two evaluator models. We find that text-based personas score the highest in almost all criteria in almost all models, consistently improving the linguistic habits of the persona in all models by a minimum $~0.2$ increase in the score. This shows that text is the most preferred way to represent a persona across models, highlighting the lack of understanding of the equivalent visual information. Some notable exceptions are in the persona consistency and expected action criteria where the descriptive image modality (\ie, a descriptive text is embedded within the image), shows a significantly higher rating than the text modality in GPT-4o, GPT-4o-mini, and Pixtral models. Since these criteria are oriented more toward the actions taken by the persona instead of the generated language, we believe the models are trained to specifically attend to embedded text to generate directed responses based on the text embedded within the image. We also note that the image and assisted image modalities consistently show similar and lower performance than others, showing that the assisted text fails to encode additional information that the models cannot already derive from the image. 

\noindent Next, we can also note that GPT-4o shows the highest average alignment scores across 4 criteria in each modality representation. This shows that GPT-4o is the most capable model of embodying these personas for the curated questions. In addition, despite the small size, we find that Pixtral is much better at capturing visual information than larger Llama models. On the other hand, Llama models shine when the persona is represented in text, significantly outperforming Pixtral. 


\paragraph{Preference-based Evaluation}
We also leverage GPT-4o as a judge to pick and choose the most aligned response with the persona, directly comparing the 4 modalities. Table~\ref{tab:voting-patterns} shows the percentage of times that the judge picks each modality in the Swiss comparison and the pairwise (only text and image) comparison setting. We find that text-based persona responses are picked for at least $90\%$ of the questions, showing a clear preference for responses generated through text-based embodiment of these models. Furthermore, in a more direct pairwise comparison, we find that image-based personas are almost never chosen (selecting text up to $99\%$). This further strengthens our claims from above regarding the lack of capabilities of current multimodal models to embody visual personas. 

\paragraph{Linguistic Evaluation}
Next, we compare the linguistic diversity of the responses generated through different persona modalities. Table~\ref{tab:ling-metrics} shows the mean and standard deviation in the three metrics of linguistic diversity in different settings. We find that text modality is the overall preferred way to generate expressive responses of a persona. Specifically, text-based personas generate at least $\sim 40$ more types of words than the other modalities, which also show significantly more variation (at least $\sim 2$ more). We note that the Llama models are highly selective and show extremely high linguistic diversity in text modality than other modalities, as Llama-3.2-90B generates up to $150$ more types when prompted with a textual persona as compared to an image and up to $2$ times root token-type ratio.

\paragraph{Human Evaluation}

We also employ independent human annotators to judge the responses generated using different persona modalities. In particular, we use GPT-4o responses of all $4$ modality representations for $10$ randomly selected questions. Each participant is shown 10 questions with a response from one of the four modalities and is asked to judge how well the response is aligned with the persona for the given question. Table~\ref{tab:human_scores} shows the mean and standard deviation of alignment scores from this study conducted on $9$ high-quality annotators. We find that our results from LLM-based evaluators are aligned with independent human annotation, showing the highest alignment for text followed by the descriptive image modality, while assisted image and image perform similarly. We defer other survey details to Appendix~\ref{app:human}.

\begin{table}[t]
    \centering
    \caption{\textbf{Human-judged alignment scores [1-4] of GPT-4o responses from different persona modalities.}}
    \label{tab:human_scores}
    \resizebox{0.7\linewidth}{!}{
    \begin{tabular}{cccc}
    \toprule
    Text  & \textbf{3.25} $\pm$ 0.91  \\
    Descriptive Image & 2.84 $\pm$ 0.98 \\
    Assisted Image &  2.71 $\pm$ 1.19 \\
    Image & 2.75 $\pm$ 1.11 \\
    \bottomrule
    \end{tabular}}
\end{table}

\subsection{Analysis on confounding factors}

\begin{figure*}[t]
    \centering
    \hspace*{\fill}
    \subfloat[Question type]{\label{fig:question} \includegraphics[width=0.19\linewidth]{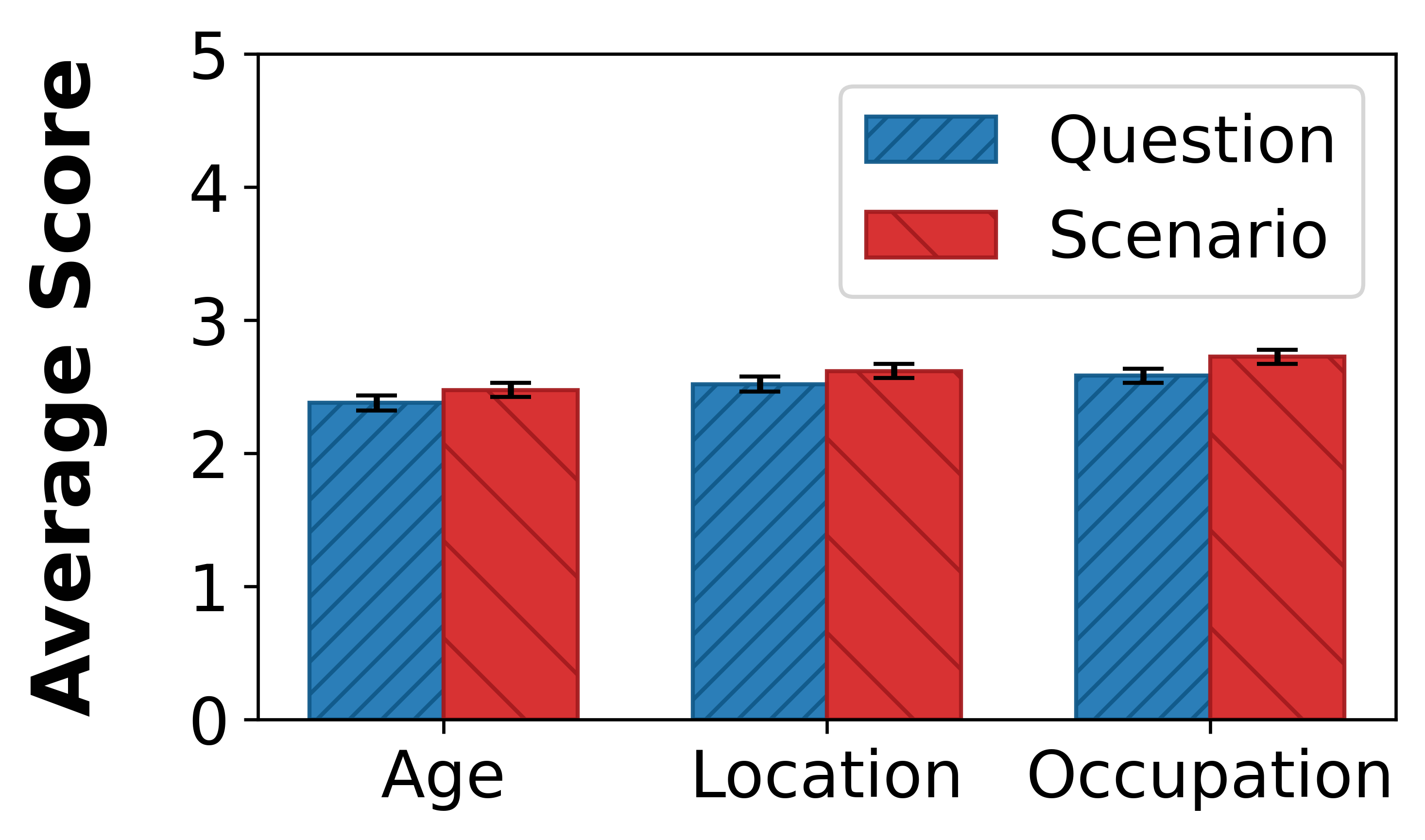}}
    \subfloat[Location]{\label{fig:location}\includegraphics[width=0.19\linewidth]{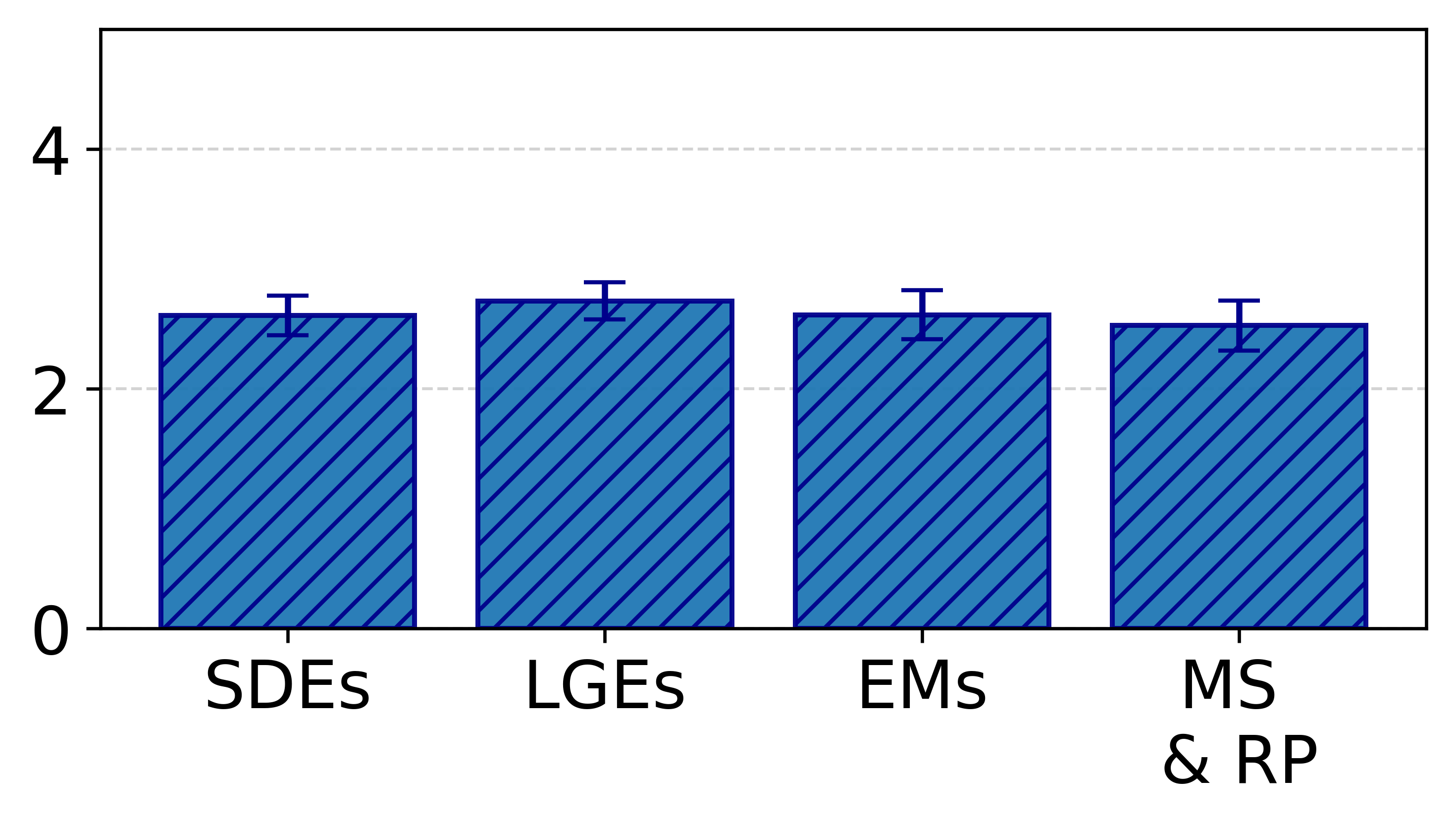}} \hfill
    \subfloat[Occupation]{\label{fig:occupation}\includegraphics[width=0.19\linewidth]{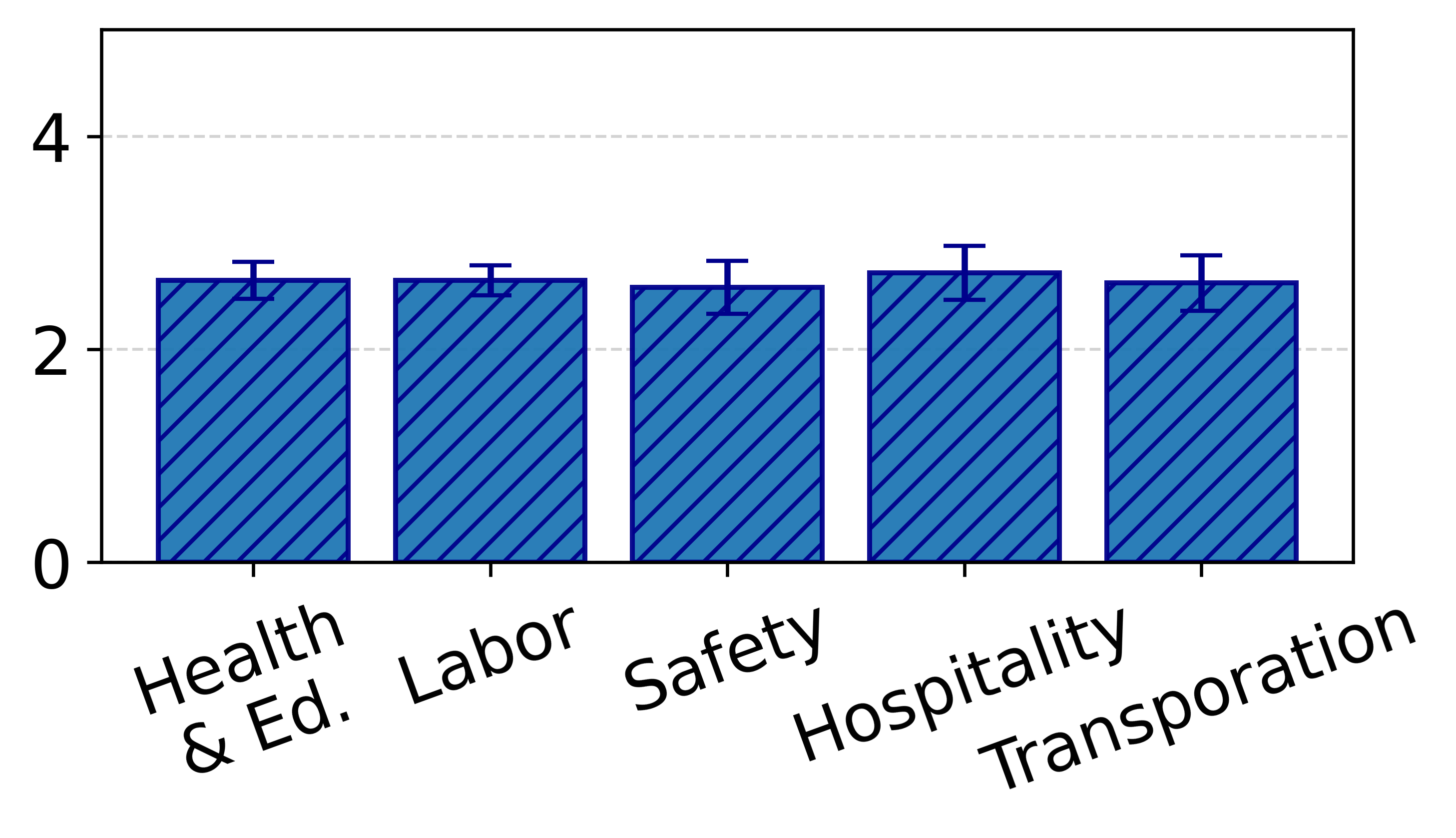}} \hfill
    \subfloat[Age]{\label{fig:age}\includegraphics[width=0.19\linewidth]{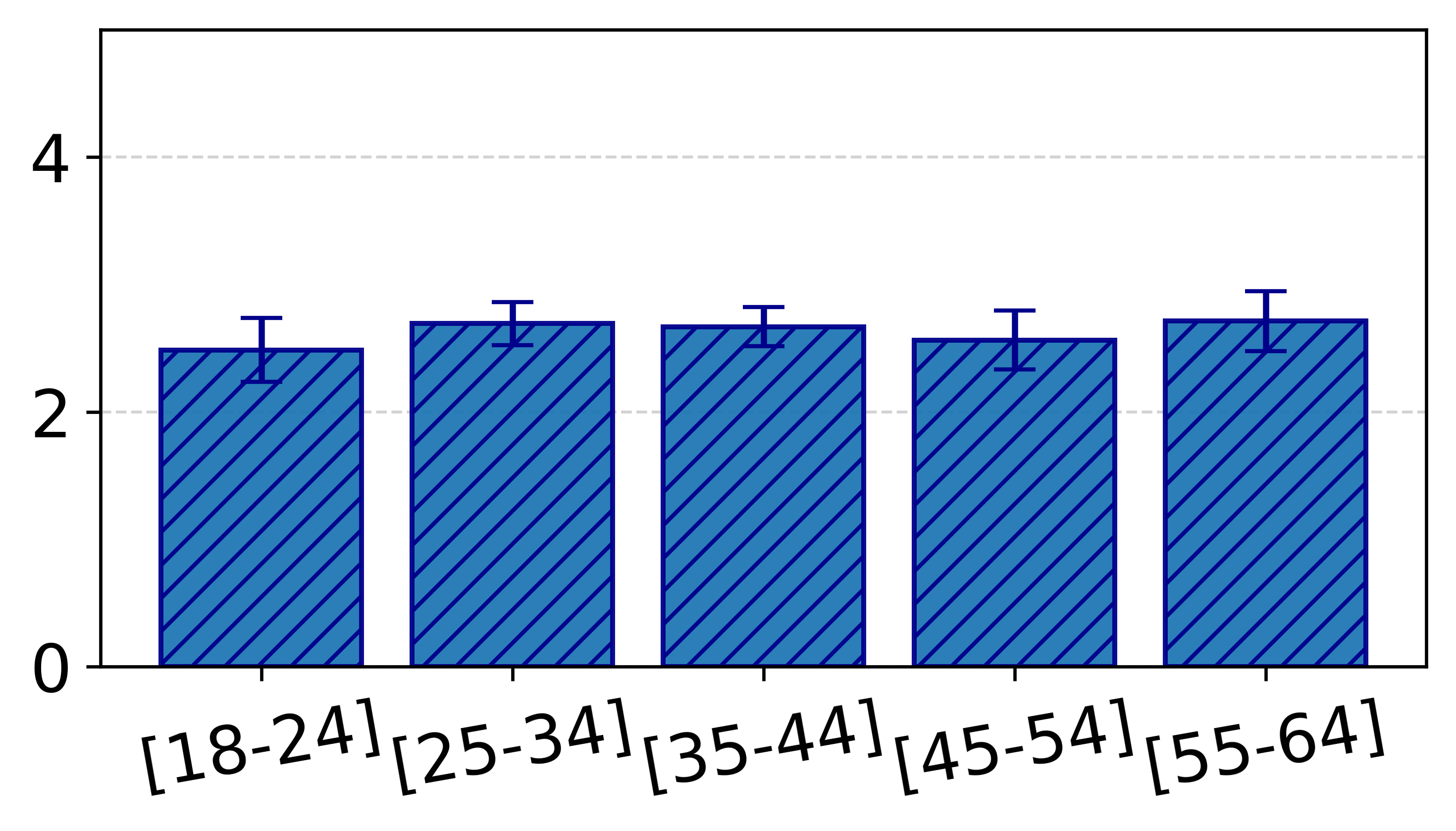}} \hfill
    \subfloat[Gender]{\label{fig:occupation}\includegraphics[width=0.19\linewidth]{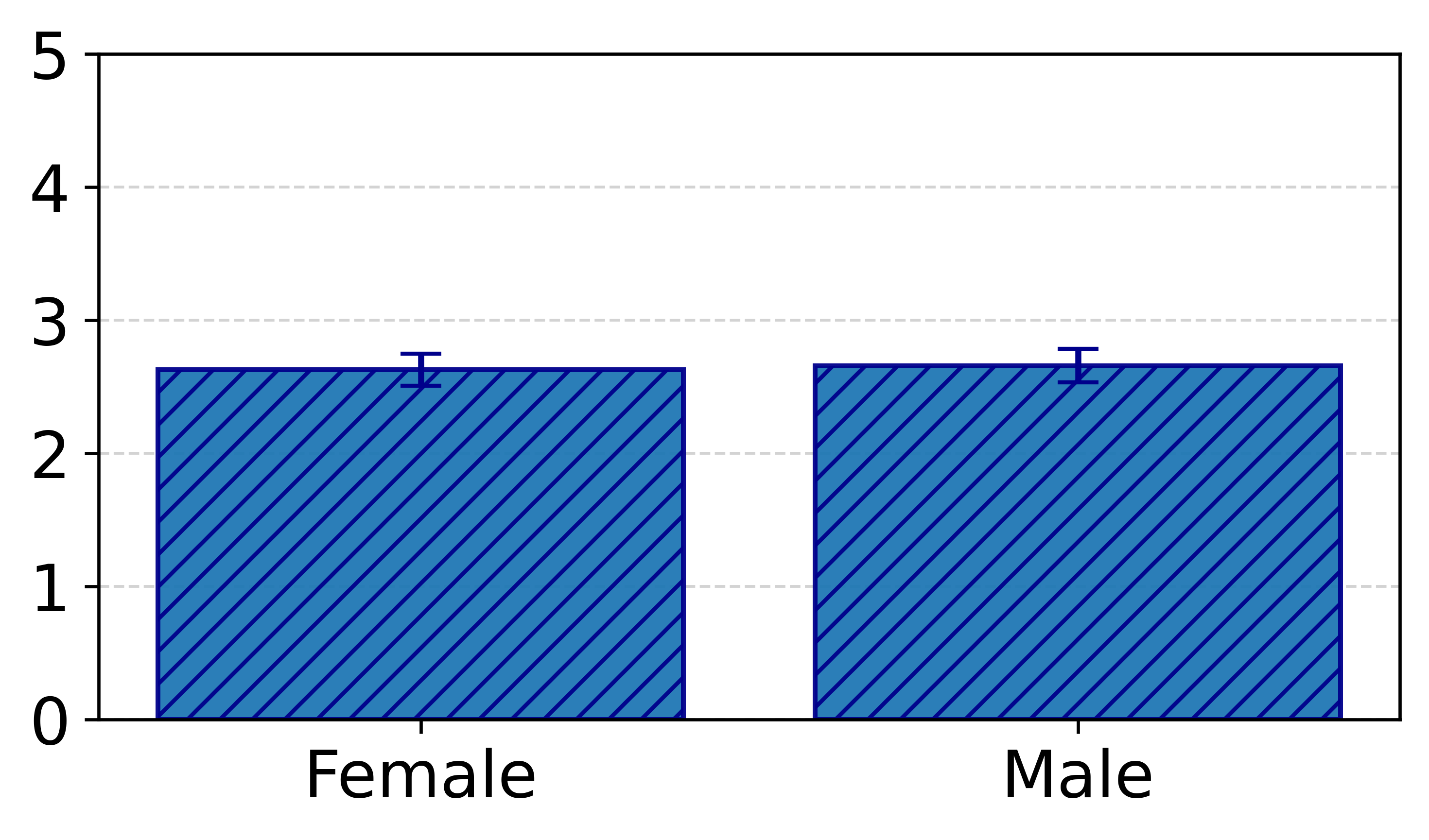}}
    \hspace*{\fill}
    \caption{LLM-based evaluation stratified based on question and persona types.}
    \label{fig:factors}
\end{figure*}

We analyze the effect of other factors that may confound our findings by stratifying the results of Table~\ref{tab:main-results-table} based on question type and attributes of the persona. Figure~\ref{fig:factors} shows the average scores assigned by LLM judge regardless of the persona modality for different categories. 
Except for a slight preference for ``scenario'' over direct ``questions'', we do not observe any major effect of these factors, confirming that the LLM-assigned scores are not confounded on factors such as question and persona attributes. This further emphasizes the role played by persona modality in Table~\ref{tab:main-results-table}. 

Finally, we also study if the results are confounded by biases in the evaluator itself, for which we compare the evaluator scores found using GPT-4o and Gemini-Flash. Tables~\ref{tab:eval-table-gpt-4o} and ~\ref{tab:eval-table-gemini-flash} in Appendix show that the trends of modality choice remain stable across the choice of these evaluators.

\section{Discussion and Related Work}\label{sec:relatedWork}

\paragraph{Persona Evaluation}

Prior work has established several frameworks for evaluating language models' role-playing capabilities. \citet{wang-etal-2024-rolellm} introduced RoleBench, an evaluation benchmark with QA pairs based on character profiles. \citet{wang-etal-2024-incharacter} developed InCharacter, assessing role-playing fidelity through psychological scales in an interview format. \citet{tu-etal-2024-charactereval} created CharacterEval, a Chinese benchmark derived from novels and scripts with multi-interaction dialogues, while \citet{shen2023roleeval} established RoleEval, a bilingual benchmark with multiple-choice questions testing persona knowledge and reasoning. \citet{samuel2024personagym} introduced PersonaGym, a dynamic evaluation framework for automated assessment of persona adherence across diverse interactions. Our work further extends the literature by performing the first systematic evaluation to understand the influence of the persona modality.

\paragraph{Multimodal Personas}

Recent work has explored integrating visual elements into LLM persona systems. \citet{ahn-etal-2023-mpchat} introduced MPCHAT, demonstrating that incorporating visual episodic memories alongside text improves dialogue consistency and persona grounding. \citet{sun-etal-2024-kiss} investigated how visual personas influence LLMs' behavior in negotiation contexts, showing models can adapt their responses based on perceived visual personality traits. \citet{dai2025mmrole} developed MMRole, a framework for training and evaluating multimodal role-playing agents. While these works establish the potential of visual personas and others extensively evaluate textual personas \citep{li-etal-2016-persona, xiao2024farllmsbelievableai, samuel2024personagym}, there has been no systematic comparison of how different modalities of persona representation affect model performance. Our work addresses this gap by directly evaluating text, visual, and hybrid approaches across a range of persona-based tasks.

\paragraph{Modality Alignment}

Language models demonstrate strong in-context learning capabilities in unimodal textual settings \citep{Shanahan2023, 10.5555/3666122.3669274}.
However, extending these capabilities to multimodal inputs remains challenging.
When visual information is introduced, models often struggle to transfer knowledge effectively from text to vision (and vice versa), resulting in noticeably weaker performance with visual in-context demonstrations compared to textual ones \citep{zhao2024mmicl, jiang2024manyshot}.
Such cross-modal gaps manifest in several ways: for instance, catastrophic forgetting of text-based instruction following can occur when models are finetuned on images \citep{zhang2024wingslearningmultimodalllms}.
While incorporating visual knowledge can yield improvements on specific tasks \citep{jin-etal-2022-leveraging}, maintaining consistently high performance across both textual and visual modalities remains an open research question, which is also highlighted in our work.
\section{Conclusion}
In this work, we study the influence of the modality of a persona in multimodal LLMs. 
We first create a novel modality-parallel dataset with equivalent representations across 4 different modality representations: image, text, assisted image, and descriptive image. 
Using a set of manually curated hard questions about the persona, we evaluate how well multimodal LLMs can represent them across different representations. 
We find a clear preference for text-based personas across $5$ multimodal LLMs, highlighting the gaps in the vision-understanding capabilities of these models in embodying diverse personas. 
Our results emphasize a grave need in the community to push the vision frontier following the advancements in language modeling. 
Given the rich amount of information that can be captured within an image, we believe it is imminent that LLM-based agents can also incorporate multimodal descriptions of their desired persona. 
We believe our modality-parallel dataset lays the foundation for future advancements in visual persona understanding in multimodal LLMs. 
Furthermore, we hope our evaluation methodology helps standardize how personas are evaluated while inspiring other evaluations in studying the influence of how the inputs are represented in LLMs.

\clearpage
\section*{Limitations}\label{sec-limitations}
A limitation of our work is that we only deal with 40 personas. However, due to a lack of any persona dataset with equivalent representations in different modalities, we see this as our contribution and leave it for future works to expand the scale of the study. Furthermore, we specifically increase the diversity of these personas across $4$ well-grounded categories, focusing on the quality of our dataset. As the field of persona alignment in LLMs is still quite nascent, we believe quality becomes more important than quantity. Additionally, it should be noted that the persona modality representations may not align perfectly across all details. Our pipeline employs two distinct mapping functions---Stable Diffusion (text-to-image) and GPT-4o-mini (image-to-text)---which will naturally introduce extraneous information or inconsistencies between representations. However, this limitation is acceptable for our evaluation framework since we only test for the presence and consistency of specific attributes rather than complete fidelity across all possible persona characteristics. Another limitation is that we have only validated our results on a small set of human annotators. We circumvent this by leveraging the validation of LLM-based evaluation with human evaluations~\citep{samuel2024personagym} while also showing a high correlation of our results across different LLM evaluators. 

\section*{Broader implications and social impact}
We intend our proposed dataset to be used strictly for academic purposes. While we design our dataset such that it does not contain any harmful and private content, our pipeline can be adapted to generate such unintended visual personas. However, we note that this is not a direct result of our artifact and can also be possible through directly querying the StableDiffusion APIs. Thus, we expect our contributions of dataset and evaluation methodology to have an overall positive social impact by inspiring future research on aligning modalities for persona embodiment.
\bibliography{citations}

\begin{thebibliography}{33}
\providecommand{\natexlab}[1]{#1}

\bibitem[{Agrawal et~al.(2024)Agrawal, Antoniak, Hanna, Bout, Chaplot, Chudnovsky, Costa, Monicault, Garg, Gervet, Ghosh, Héliou, Jacob, Jiang, Khandelwal, Lacroix, Lample, Casas, Lavril, Scao, Lo, Marshall, Martin, Mensch, Muddireddy, Nemychnikova, Pellat, Platen, Raghuraman, Rozière, Sablayrolles, Saulnier, Sauvestre, Shang, Soletskyi, Stewart, Stock, Studnia, Subramanian, Vaze, Wang, and Yang}]{agrawal2024pixtral12b}
Pravesh Agrawal, Szymon Antoniak, Emma~Bou Hanna, Baptiste Bout, Devendra Chaplot, Jessica Chudnovsky, Diogo Costa, Baudouin~De Monicault, Saurabh Garg, Theophile Gervet, Soham Ghosh, Amélie Héliou, Paul Jacob, Albert~Q. Jiang, Kartik Khandelwal, Timothée Lacroix, Guillaume Lample, Diego~Las Casas, Thibaut Lavril, and 23 others. 2024.
\newblock \href {https://arxiv.org/abs/2410.07073} {Pixtral 12b}.
\newblock \emph{Preprint}, arXiv:2410.07073.

\bibitem[{Ahn et~al.(2023)Ahn, Song, Yun, and Kim}]{ahn-etal-2023-mpchat}
Jaewoo Ahn, Yeda Song, Sangdoo Yun, and Gunhee Kim. 2023.
\newblock \href {https://doi.org/10.18653/v1/2023.acl-long.189} {{MPCHAT}: Towards multimodal persona-grounded conversation}.
\newblock In \emph{Proceedings of the 61st Annual Meeting of the Association for Computational Linguistics (Volume 1: Long Papers)}, pages 3354--3377, Toronto, Canada. Association for Computational Linguistics.

\bibitem[{Chen et~al.(2024)Chen, Wang, Tian, Ye, Gao, Cui, Tong, Hu, Luo, Ma et~al.}]{chen2024far}
Zhe Chen, Weiyun Wang, Hao Tian, Shenglong Ye, Zhangwei Gao, Erfei Cui, Wenwen Tong, Kongzhi Hu, Jiapeng Luo, Zheng Ma, and 1 others. 2024.
\newblock How far are we to gpt-4v? closing the gap to commercial multimodal models with open-source suites.
\newblock \emph{Science China Information Sciences}, 67(12):220101.

\bibitem[{Dai et~al.(2025)Dai, Hu, Wang, Jin, Chen, and Lu}]{dai2025mmrole}
Yanqi Dai, Huanran Hu, Lei Wang, Shengjie Jin, Xu~Chen, and Zhiwu Lu. 2025.
\newblock \href {https://openreview.net/forum?id=FGSgsefE0Y} {{MMR}ole: A comprehensive framework for developing and evaluating multimodal role-playing agents}.
\newblock In \emph{The Thirteenth International Conference on Learning Representations}.

\bibitem[{GPT-4o(2025)}]{gpt4o}
GPT-4o. 2025.
\newblock \url{https://openai.com/index/hello-gpt-4o/}.
\newblock [Accessed 14-02-2025].

\bibitem[{Grattafiori et~al.(2024)Grattafiori, Dubey, Jauhri, Pandey, Kadian, Al-Dahle, Letman, Mathur, Schelten, Vaughan, Yang, Fan, Goyal, Hartshorn, Yang, Mitra, Sravankumar, Korenev, Hinsvark, Rao, Zhang, Rodriguez, Gregerson, Spataru, Roziere, Biron, Tang, Chern, Caucheteux, Nayak, Bi, Marra, McConnell, Keller, Touret, Wu, Wong, Ferrer, Nikolaidis, Allonsius, Song, Pintz, Livshits, Wyatt, Esiobu, Choudhary, Mahajan, Garcia-Olano, Perino, Hupkes, Lakomkin, AlBadawy, Lobanova, Dinan, Smith, Radenovic, Guzmán, Zhang, Synnaeve, Lee, Anderson, Thattai, Nail, Mialon, Pang, Cucurell, Nguyen, Korevaar, Xu, Touvron, Zarov, Ibarra, Kloumann, Misra, Evtimov, Zhang, Copet, Lee, Geffert, Vranes, Park, Mahadeokar, Shah, van~der Linde, Billock, Hong, Lee, Fu, Chi, Huang, Liu, Wang, Yu, Bitton, Spisak, Park, Rocca, Johnstun, Saxe, Jia, Alwala, Prasad, Upasani, Plawiak, Li, Heafield, Stone, El-Arini, Iyer, Malik, Chiu, Bhalla, Lakhotia, Rantala-Yeary, van~der Maaten, Chen, Tan, Jenkins, Martin, Madaan, Malo, Blecher,
  Landzaat, de~Oliveira, Muzzi, Pasupuleti, Singh, Paluri, Kardas, Tsimpoukelli, Oldham, Rita, Pavlova, Kambadur, Lewis, Si, Singh, Hassan, Goyal, Torabi, Bashlykov, Bogoychev, Chatterji, Zhang, Duchenne, Çelebi, Alrassy, Zhang, Li, Vasic, Weng, Bhargava, Dubal, Krishnan, Koura, Xu, He, Dong, Srinivasan, Ganapathy, Calderer, Cabral, Stojnic, Raileanu, Maheswari, Girdhar, Patel, Sauvestre, Polidoro, Sumbaly, Taylor, Silva, Hou, Wang, Hosseini, Chennabasappa, Singh, Bell, Kim, Edunov, Nie, Narang, Raparthy, Shen, Wan, Bhosale, Zhang, Vandenhende, Batra, Whitman, Sootla, Collot, Gururangan, Borodinsky, Herman, Fowler, Sheasha, Georgiou, Scialom, Speckbacher, Mihaylov, Xiao, Karn, Goswami, Gupta, Ramanathan, Kerkez, Gonguet, Do, Vogeti, Albiero, Petrovic, Chu, Xiong, Fu, Meers, Martinet, Wang, Wang, Tan, Xia, Xie, Jia, Wang, Goldschlag, Gaur, Babaei, Wen, Song, Zhang, Li, Mao, Coudert, Yan, Chen, Papakipos, Singh, Srivastava, Jain, Kelsey, Shajnfeld, Gangidi, Victoria, Goldstand, Menon, Sharma, Boesenberg,
  Baevski, Feinstein, Kallet, Sangani, Teo, Yunus, Lupu, Alvarado, Caples, Gu, Ho, Poulton, Ryan, Ramchandani, Dong, Franco, Goyal, Saraf, Chowdhury, Gabriel, Bharambe, Eisenman, Yazdan, James, Maurer, Leonhardi, Huang, Loyd, Paola, Paranjape, Liu, Wu, Ni, Hancock, Wasti, Spence, Stojkovic, Gamido, Montalvo, Parker, Burton, Mejia, Liu, Wang, Kim, Zhou, Hu, Chu, Cai, Tindal, Feichtenhofer, Gao, Civin, Beaty, Kreymer, Li, Adkins, Xu, Testuggine, David, Parikh, Liskovich, Foss, Wang, Le, Holland, Dowling, Jamil, Montgomery, Presani, Hahn, Wood, Le, Brinkman, Arcaute, Dunbar, Smothers, Sun, Kreuk, Tian, Kokkinos, Ozgenel, Caggioni, Kanayet, Seide, Florez, Schwarz, Badeer, Swee, Halpern, Herman, Sizov, Guangyi, Zhang, Lakshminarayanan, Inan, Shojanazeri, Zou, Wang, Zha, Habeeb, Rudolph, Suk, Aspegren, Goldman, Zhan, Damlaj, Molybog, Tufanov, Leontiadis, Veliche, Gat, Weissman, Geboski, Kohli, Lam, Asher, Gaya, Marcus, Tang, Chan, Zhen, Reizenstein, Teboul, Zhong, Jin, Yang, Cummings, Carvill, Shepard, McPhie,
  Torres, Ginsburg, Wang, Wu, U, Saxena, Khandelwal, Zand, Matosich, Veeraraghavan, Michelena, Li, Jagadeesh, Huang, Chawla, Huang, Chen, Garg, A, Silva, Bell, Zhang, Guo, Yu, Moshkovich, Wehrstedt, Khabsa, Avalani, Bhatt, Mankus, Hasson, Lennie, Reso, Groshev, Naumov, Lathi, Keneally, Liu, Seltzer, Valko, Restrepo, Patel, Vyatskov, Samvelyan, Clark, Macey, Wang, Hermoso, Metanat, Rastegari, Bansal, Santhanam, Parks, White, Bawa, Singhal, Egebo, Usunier, Mehta, Laptev, Dong, Cheng, Chernoguz, Hart, Salpekar, Kalinli, Kent, Parekh, Saab, Balaji, Rittner, Bontrager, Roux, Dollar, Zvyagina, Ratanchandani, Yuvraj, Liang, Alao, Rodriguez, Ayub, Murthy, Nayani, Mitra, Parthasarathy, Li, Hogan, Battey, Wang, Howes, Rinott, Mehta, Siby, Bondu, Datta, Chugh, Hunt, Dhillon, Sidorov, Pan, Mahajan, Verma, Yamamoto, Ramaswamy, Lindsay, Lindsay, Feng, Lin, Zha, Patil, Shankar, Zhang, Zhang, Wang, Agarwal, Sajuyigbe, Chintala, Max, Chen, Kehoe, Satterfield, Govindaprasad, Gupta, Deng, Cho, Virk, Subramanian, Choudhury,
  Goldman, Remez, Glaser, Best, Koehler, Robinson, Li, Zhang, Matthews, Chou, Shaked, Vontimitta, Ajayi, Montanez, Mohan, Kumar, Mangla, Ionescu, Poenaru, Mihailescu, Ivanov, Li, Wang, Jiang, Bouaziz, Constable, Tang, Wu, Wang, Wu, Gao, Kleinman, Chen, Hu, Jia, Qi, Li, Zhang, Zhang, Adi, Nam, Yu, Wang, Zhao, Hao, Qian, Li, He, Rait, DeVito, Rosnbrick, Wen, Yang, Zhao, and Ma}]{grattafiori2024llama3herdmodels}
Aaron Grattafiori, Abhimanyu Dubey, Abhinav Jauhri, Abhinav Pandey, Abhishek Kadian, Ahmad Al-Dahle, Aiesha Letman, Akhil Mathur, Alan Schelten, Alex Vaughan, Amy Yang, Angela Fan, Anirudh Goyal, Anthony Hartshorn, Aobo Yang, Archi Mitra, Archie Sravankumar, Artem Korenev, Arthur Hinsvark, and 542 others. 2024.
\newblock \href {https://arxiv.org/abs/2407.21783} {The llama 3 herd of models}.
\newblock \emph{Preprint}, arXiv:2407.21783.

\bibitem[{Jiang et~al.(2024)Jiang, Irvin, Wang, Chaudhry, Chen, and Ng}]{jiang2024manyshot}
Yixing Jiang, Jeremy~Andrew Irvin, Ji~Hun Wang, Muhammad~Ahmed Chaudhry, Jonathan~H Chen, and Andrew~Y. Ng. 2024.
\newblock \href {https://openreview.net/forum?id=j2rKwWXdcz} {Many-shot in-context learning in multimodal foundation models}.
\newblock In \emph{ICML 2024 Workshop on In-Context Learning}.

\bibitem[{Jin et~al.(2022)Jin, Lee, Zhu, Pujara, and Ren}]{jin-etal-2022-leveraging}
Woojeong Jin, Dong-Ho Lee, Chenguang Zhu, Jay Pujara, and Xiang Ren. 2022.
\newblock \href {https://doi.org/10.18653/v1/2022.acl-long.196} {Leveraging visual knowledge in language tasks: An empirical study on intermediate pre-training for cross-modal knowledge transfer}.
\newblock In \emph{Proceedings of the 60th Annual Meeting of the Association for Computational Linguistics (Volume 1: Long Papers)}, pages 2750--2762, Dublin, Ireland. Association for Computational Linguistics.

\bibitem[{Li et~al.(2024{\natexlab{a}})Li, Wong, Zhang, Usuyama, Liu, Yang, Naumann, Poon, and Gao}]{li2024llavamed}
Chunyuan Li, Cliff Wong, Sheng Zhang, Naoto Usuyama, Haotian Liu, Jianwei Yang, Tristan Naumann, Hoifung Poon, and Jianfeng Gao. 2024{\natexlab{a}}.
\newblock Llava-med: Training a large language-and-vision assistant for biomedicine in one day.
\newblock \emph{Advances in Neural Information Processing Systems}, 36.

\bibitem[{Li et~al.(2016)Li, Galley, Brockett, Spithourakis, Gao, and Dolan}]{li-etal-2016-persona}
Jiwei Li, Michel Galley, Chris Brockett, Georgios Spithourakis, Jianfeng Gao, and Bill Dolan. 2016.
\newblock \href {https://doi.org/10.18653/v1/P16-1094} {A persona-based neural conversation model}.
\newblock In \emph{Proceedings of the 54th Annual Meeting of the Association for Computational Linguistics (Volume 1: Long Papers)}, pages 994--1003, Berlin, Germany. Association for Computational Linguistics.

\bibitem[{Li et~al.(2024{\natexlab{b}})Li, Wen, Wang, Li, Yuan, Liu, Liu, Xu, Wang, Sun et~al.}]{li2024personal}
Yuanchun Li, Hao Wen, Weijun Wang, Xiangyu Li, Yizhen Yuan, Guohong Liu, Jiacheng Liu, Wenxing Xu, Xiang Wang, Yi~Sun, and 1 others. 2024{\natexlab{b}}.
\newblock Personal llm agents: Insights and survey about the capability, efficiency and security.
\newblock \emph{arXiv preprint arXiv:2401.05459}.

\bibitem[{Liu et~al.(2024)Liu, Li, Wu, and Lee}]{liu2024visual}
Haotian Liu, Chunyuan Li, Qingyang Wu, and Yong~Jae Lee. 2024.
\newblock Visual instruction tuning.
\newblock \emph{Advances in neural information processing systems}, 36.

\bibitem[{Ma et~al.(2024)Ma, Luo, Wang, Liu, Chen, Li, and Xiao}]{ma2024visual}
Siyuan Ma, Weidi Luo, Yu~Wang, Xiaogeng Liu, Muhao Chen, Bo~Li, and Chaowei Xiao. 2024.
\newblock Visual-roleplay: Universal jailbreak attack on multimodal large language models via role-playing image characte.
\newblock \emph{arXiv preprint arXiv:2405.20773}.

\bibitem[{McCarthy and Jarvis(2010)}]{mccarthy2010mtld}
Philip~M McCarthy and Scott Jarvis. 2010.
\newblock Mtld, vocd-d, and hd-d: A validation study of sophisticated approaches to lexical diversity assessment.
\newblock \emph{Behavior research methods}, 42(2):381--392.

\bibitem[{Paivio(2013)}]{paivio2013imagery}
Allan Paivio. 2013.
\newblock \emph{Imagery and verbal processes}.
\newblock Psychology Press.

\bibitem[{ProtectAI.com(2024)}]{distilroberta-base-rejection-v1}
ProtectAI.com. 2024.
\newblock \href {https://huggingface.co/ProtectAI/distilroberta-base-rejection-v1} {Fine-tuned distilroberta-base for rejection in the output detection}.

\bibitem[{Salewski et~al.(2023)Salewski, Alaniz, Rio-Torto, Schulz, and Akata}]{10.5555/3666122.3669274}
Leonard Salewski, Stephan Alaniz, Isabel Rio-Torto, Eric Schulz, and Zeynep Akata. 2023.
\newblock In-context impersonation reveals large language models' strengths and biases.
\newblock In \emph{Proceedings of the 37th International Conference on Neural Information Processing Systems}, NIPS '23, Red Hook, NY, USA. Curran Associates Inc.

\bibitem[{Samuel et~al.(2024)Samuel, Zou, Zhou, Chaudhari, Kalyan, Rajpurohit, Deshpande, Narasimhan, and Murahari}]{samuel2024personagym}
Vinay Samuel, Henry~Peng Zou, Yue Zhou, Shreyas Chaudhari, Ashwin Kalyan, Tanmay Rajpurohit, Ameet Deshpande, Karthik Narasimhan, and Vishvak Murahari. 2024.
\newblock Personagym: Evaluating persona agents and llms.
\newblock \emph{arXiv preprint arXiv:2407.18416}.

\bibitem[{Shanahan et~al.(2023)Shanahan, McDonell, and Reynolds}]{Shanahan2023}
Murray Shanahan, K.~McDonell, and L.~Reynolds. 2023.
\newblock \href {https://doi.org/10.1038/s41586-023-06647-8} {Role play with large language models}.
\newblock \emph{Nature}, 623:493--498.

\bibitem[{Shen et~al.(2023)Shen, Li, and Xiong}]{shen2023roleeval}
Tianhao Shen, Sun Li, and Deyi Xiong. 2023.
\newblock \href {https://arxiv.org/abs/2312.16132} {Roleeval: A bilingual role evaluation benchmark for large language models}.

\bibitem[{Sun et~al.(2024)Sun, Lee, Baek, Hwang, Lee, Nan, Jansen, and Kim}]{sun-etal-2024-kiss}
Seungjong Sun, Eungu Lee, Seo~Yeon Baek, Seunghyun Hwang, Wonbyung Lee, Dongyan Nan, Bernard~J Jansen, and Jang~Hyun Kim. 2024.
\newblock \href {https://doi.org/10.18653/v1/2024.emnlp-main.609} {Kiss up, kick down: Exploring behavioral changes in multi-modal large language models with assigned visual personas}.
\newblock In \emph{Proceedings of the 2024 Conference on Empirical Methods in Natural Language Processing}, pages 10888--10901, Miami, Florida, USA. Association for Computational Linguistics.

\bibitem[{Todorov et~al.(2015)Todorov, Olivola, Dotsch, and Mende-Siedlecki}]{todorov2015social}
Alexander Todorov, Christopher~Y Olivola, Ron Dotsch, and Peter Mende-Siedlecki. 2015.
\newblock Social attributions from faces: Determinants, consequences, accuracy, and functional significance.
\newblock \emph{Annual review of psychology}, 66(1):519--545.

\bibitem[{Tong et~al.(2024)Tong, Liu, Zhai, Ma, LeCun, and Xie}]{tong2024eyes}
Shengbang Tong, Zhuang Liu, Yuexiang Zhai, Yi~Ma, Yann LeCun, and Saining Xie. 2024.
\newblock Eyes wide shut? exploring the visual shortcomings of multimodal llms.
\newblock In \emph{Proceedings of the IEEE/CVF Conference on Computer Vision and Pattern Recognition}, pages 9568--9578.

\bibitem[{Tseng et~al.(2024)Tseng, Huang, Hsiao, Hsu, Foo, Huang, and Chen}]{tseng2024two}
Yu-Min Tseng, Yu-Chao Huang, Teng-Yun Hsiao, Yu-Ching Hsu, Jia-Yin Foo, Chao-Wei Huang, and Yun-Nung Chen. 2024.
\newblock Two tales of persona in llms: A survey of role-playing and personalization.
\newblock \emph{arXiv preprint arXiv:2406.01171}.

\bibitem[{Tu et~al.(2024)Tu, Fan, Tian, Shen, Shang, Gao, and Yan}]{tu-etal-2024-charactereval}
Quan Tu, Shilong Fan, Zihang Tian, Tianhao Shen, Shuo Shang, Xin Gao, and Rui Yan. 2024.
\newblock \href {https://doi.org/10.18653/v1/2024.acl-long.638} {{C}haracter{E}val: A {C}hinese benchmark for role-playing conversational agent evaluation}.
\newblock In \emph{Proceedings of the 62nd Annual Meeting of the Association for Computational Linguistics (Volume 1: Long Papers)}, pages 11836--11850, Bangkok, Thailand. Association for Computational Linguistics.

\bibitem[{{van Hout} and Vermeer(2007)}]{lexicalrichnesshout}
R.~{van Hout} and A.~Vermeer. 2007.
\newblock \emph{Comparing measures of lexical richness}, pages 93--116.
\newblock Cambridge University Press, United Kingdom.
\newblock Pagination: 23.

\bibitem[{Wang et~al.(2024{\natexlab{a}})Wang, Peng, Que, Liu, Zhou, Wu, Guo, Gan, Ni, Yang, Zhang, Zhang, Ouyang, Xu, Huang, Fu, and Peng}]{wang-etal-2024-rolellm}
Noah Wang, Z.y. Peng, Haoran Que, Jiaheng Liu, Wangchunshu Zhou, Yuhan Wu, Hongcheng Guo, Ruitong Gan, Zehao Ni, Jian Yang, Man Zhang, Zhaoxiang Zhang, Wanli Ouyang, Ke~Xu, Wenhao Huang, Jie Fu, and Junran Peng. 2024{\natexlab{a}}.
\newblock \href {https://doi.org/10.18653/v1/2024.findings-acl.878} {{R}ole{LLM}: Benchmarking, eliciting, and enhancing role-playing abilities of large language models}.
\newblock In \emph{Findings of the Association for Computational Linguistics: ACL 2024}, pages 14743--14777, Bangkok, Thailand. Association for Computational Linguistics.

\bibitem[{Wang et~al.(2024{\natexlab{b}})Wang, Xiao, Huang, Yuan, Xu, Guo, Tu, Fei, Leng, Wang, Chen, Li, and Xiao}]{wang-etal-2024-incharacter}
Xintao Wang, Yunze Xiao, Jen-tse Huang, Siyu Yuan, Rui Xu, Haoran Guo, Quan Tu, Yaying Fei, Ziang Leng, Wei Wang, Jiangjie Chen, Cheng Li, and Yanghua Xiao. 2024{\natexlab{b}}.
\newblock \href {https://doi.org/10.18653/v1/2024.acl-long.102} {{I}n{C}haracter: Evaluating personality fidelity in role-playing agents through psychological interviews}.
\newblock In \emph{Proceedings of the 62nd Annual Meeting of the Association for Computational Linguistics (Volume 1: Long Papers)}, pages 1840--1873, Bangkok, Thailand. Association for Computational Linguistics.

\bibitem[{Wei et~al.(2024)Wei, Haghtalab, and Steinhardt}]{wei2024jailbroken}
Alexander Wei, Nika Haghtalab, and Jacob Steinhardt. 2024.
\newblock Jailbroken: How does llm safety training fail?
\newblock \emph{Advances in Neural Information Processing Systems}, 36.

\bibitem[{Xiao et~al.(2024)Xiao, Cheng, Fu, Wang, Li, and Liu}]{xiao2024farllmsbelievableai}
Yang Xiao, Yi~Cheng, Jinlan Fu, Jiashuo Wang, Wenjie Li, and Pengfei Liu. 2024.
\newblock \href {https://arxiv.org/abs/2312.17115} {How far are llms from believable ai? a benchmark for evaluating the believability of human behavior simulation}.
\newblock \emph{Preprint}, arXiv:2312.17115.

\bibitem[{Zhang et~al.(2024)Zhang, Lu, Li, Ma, Chen, Xu, Luo, Zhang, Zhan, and Ye}]{zhang2024wingslearningmultimodalllms}
Yi-Kai Zhang, Shiyin Lu, Yang Li, Yanqing Ma, Qing-Guo Chen, Zhao Xu, Weihua Luo, Kaifu Zhang, De-Chuan Zhan, and Han-Jia Ye. 2024.
\newblock \href {https://arxiv.org/abs/2406.03496} {Wings: Learning multimodal llms without text-only forgetting}.
\newblock \emph{Preprint}, arXiv:2406.03496.

\bibitem[{Zhao et~al.(2024{\natexlab{a}})Zhao, Cai, Si, Ma, An, Chen, Liu, Wang, Han, and Chang}]{zhao2024mmicl}
Haozhe Zhao, Zefan Cai, Shuzheng Si, Xiaojian Ma, Kaikai An, Liang Chen, Zixuan Liu, Sheng Wang, Wenjuan Han, and Baobao Chang. 2024{\natexlab{a}}.
\newblock \href {https://openreview.net/forum?id=5KojubHBr8} {{MMICL}: Empowering vision-language model with multi-modal in-context learning}.
\newblock In \emph{The Twelfth International Conference on Learning Representations}.

\bibitem[{Zhao et~al.(2024{\natexlab{b}})Zhao, Qian, Cao, Wang, and Ding}]{zhao2024bias}
Jinman Zhao, Zifan Qian, Linbo Cao, Yining Wang, and Yitian Ding. 2024{\natexlab{b}}.
\newblock Bias and toxicity in role-play reasoning.
\newblock \emph{arXiv preprint arXiv:2409.13979}.

\end{thebibliography}

\clearpage

\appendix
\appendix

\section*{Appendix}

\section{Prompts}\label{app:prompts}
\subsection{Textual Description}\label{app:img_to_text_prompt}
\begin{quote}
    {\small
    \texttt{Create a short, descriptive persona for the person in the image. Describe them using only the following details: their age, gender, facial expression or mood, attire, any tools or items they’re holding, their work environment, the nature of their job, and their connection to the area and location. Avoid taking creative liberties beyond these details, only using details that can be inferred from the image, while aiming for a realistic portrayal that gives insight into their daily life, professional dedication, and overall demeanor. For example: Meet a skilled construction worker in his late 30s, living in Sydney, Australia. Every day, he heads out to work in one of the city's bustling urban sites, often with a view of iconic landmarks like the Sydney Opera House and Sydney Harbour Bridge. Outfitted in essential safety gear—a hard hat, reflective vest, and a set of versatile tools—he’s well-prepared for a physically demanding role that demands focus and precision. His job involves a blend of construction and maintenance tasks, requiring him to pay close attention to safety protocols and collaborate with a team. Confident and professional in his work, he takes pride in contributing to the infrastructure and vibrant aesthetic of Sydney, adding to the city’s ever-evolving landscape with each project.
    }}
\end{quote}

\begin{table*}[t]
    \centering
    \caption{A complete list of personas annotated for their attribute categories.}
    \label{tab:personalist}
    \resizebox{1.0\linewidth}{!}{
    \begin{tabular}{l c c c c}
        \toprule
        Persona & Age & Gender & Occupation & Location \\
        \midrule
        A 25-year-old female nurse from Toronto & 25-34 & female & healthcare \& education & Strong Developed Economies \\ 
        A 41-year-old female electrician from Sydney & 35-44 & female & manual labor & Strong Developed Economies \\ 
        A 36-year-old male electrician from Houston & 35-44 & male & manual labor & Largest Global Economies \\ 
        A 29-year-old female police officer from New York & 25-34 & female & public safety & Largest Global Economies \\ 
        A 28-year-old female police officer from London & 25-34 & female & public safety & Largest Global Economies \\ 
        A 35-year-old male chef from Paris & 35-44 & male & hospitality & Largest Global Economies \\ 
        A 32-year-old female chef from Rome & 25-34 & female & hospitality & Strong Developed Economies \\ 
        A 50-year-old male farmer from Sao Paulo & 45-54 & male & manual labor & Emerging Markets \\ 
        A 40-year-old female farmer from Nairobi & 35-44 & female & manual labor & Emerging Markets \\ 
        A 27-year-old female mechanic from Berlin & 25-34 & female & manual labor & Largest Global Economies \\ 
        A 28-year-old female pilot from Los Angeles & 25-34 & female & transportation & Largest Global Economies \\ 
        A 28-year-old female pilot from Vancouver & 25-34 & female & transportation & Strong Developed Economies \\ 
        A 60-year-old female carpenter from Rome & 55-64 & female & manual labor & Strong Developed Economies \\ 
        A 45-year-old male carpenter from Auckland & 45-54 & male & manual labor & Emerging Markets \\ 
        A 44-year-old female cashier from Montreal & 35-44 & female & hospitality & Strong Developed Economies \\ 
        A 56-year-old male roofer from Brisbane & 55-64 & male & manual labor & Strong Developed Economies \\ 
        A 30-year-old female garbage collector from Toronto & 25-34 & female & manual labor & Strong Developed Economies \\ 
        A 63-year-old male miner from Johannesburg & 55-64 & male & manual labor & Emerging Markets \\ 
        A 24-year-old female lab technician from Shanghai & 18-24 & female & healthcare \& education & Largest Global Economies \\ 
        A 29-year-old male postal worker from Mexico City & 25-34 & male & transportation & Emerging Markets \\ 
        A 44-year-old female welder from Dubai & 35-44 & female & manual labor & Mid-Sized \& Regional Powers \\ 
        A 54-year-old male librarian from Amsterdam & 45-54 & male & healthcare \& education & Mid-Sized \& Regional Powers \\ 
        A 51-year-old female dentist from Seoul & 45-54 & female & healthcare \& education & Strong Developed Economies \\ 
        A 40-year-old female landscaper from Edinburgh & 35-44 & female & manual labor & Largest Global Economies \\ 
        A 24-year-old male hairdresser from Barcelona & 18-24 & male & hospitality & Strong Developed Economies \\ 
        A 19-year-old male janitor from Stockholm & 18-24 & male & manual labor & Mid-Sized \& Regional Powers \\ 
        A 53-year-old female bus driver from Copenhagen & 45-54 & female & transportation & Mid-Sized \& Regional Powers \\ 
        A 27-year-old female machinist from Frankfurt & 25-34 & female & manual labor & Largest Global Economies \\ 
        A 52-year-old male doctor from Madrid & 45-54 & male & healthcare \& education & Strong Developed Economies \\ 
        A 60-year-old male security guard from Lisbon & 55-64 & male & public safety & Mid-Sized \& Regional Powers \\ 
        A 42-year-old male firefighter from Sao Paulo & 35-44 & male & public safety & Emerging Markets \\ 
        A 36-year-old male pharmacist from Berlin & 35-44 & male & healthcare \& education & Largest Global Economies \\ 
        A 56-year-old female teacher from Melbourne & 55-64 & female & healthcare \& education & Strong Developed Economies \\ 
        A 42-year-old male taxi driver from Hong Kong & 35-44 & male & transportation & Largest Global Economies \\ 
        A 39-year-old female veterinarian from Nairobi & 35-44 & female & healthcare \& education & Emerging Markets \\ 
        A 25-year-old male baker from Lisbon & 25-34 & male & hospitality & Mid-Sized \& Regional Powers \\ 
        A 40-year-old male welder from Moscow & 35-44 & male & manual labor & Mid-Sized \& Regional Powers \\ 
        A 39-year-old male plumber from Melbourne & 35-44 & male & manual labor & Strong Developed Economies \\ 
        A 22-year-old male lab technician from Tokyo & 18-24 & male & healthcare \& education & Largest Global Economies \\ 
        A 20-year-old female security guard from Cape Town & 18-24 & female & public safety & Emerging Markets   \\
        \bottomrule
    \end{tabular}
    }
\end{table*}

\subsection{Effect of Safety Training}
\label{app:safety-training}

\begin{figure}[t]
        \centering
        \includegraphics[width=\linewidth]{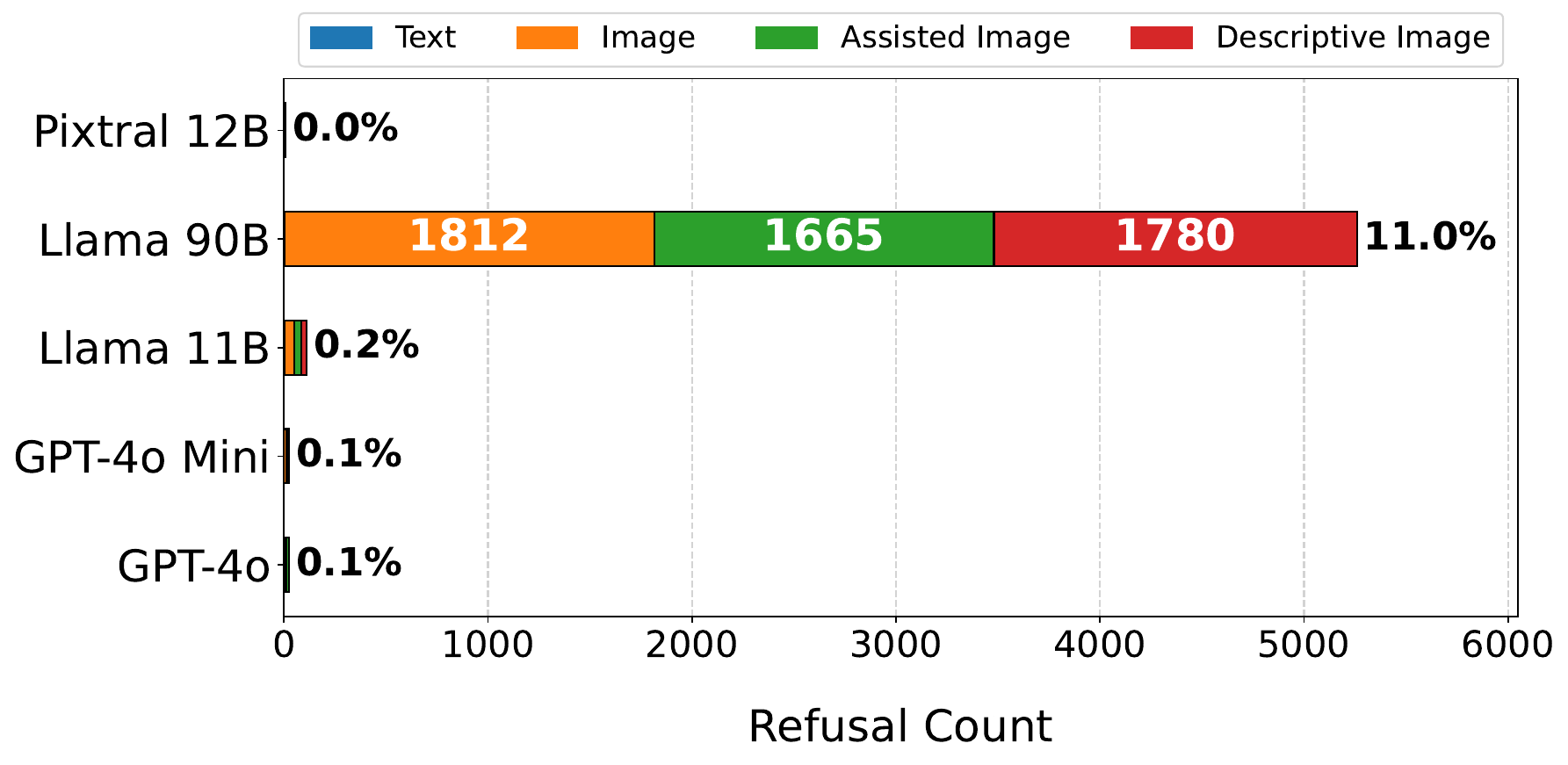}
        \caption{The rate and number of refusals in response to persona prompts. Llama 3.2 90B shows a strong aversion to multimodal persona prompts, while other models rarely refuse.}
    \label{fig:safety-training}
\end{figure}

In our experiments, we observed that Llama 3.2 90B frequently refused to assume visual personas\footnote{Refusal detection was performed using a fine-tuned \texttt{distilroberta-base} model \citep{distilroberta-base-rejection-v1}}, refusing to engage with 76.7\% of all visual persona prompts (Figure \ref{fig:safety-training}). This behavior can be attributed to an overgeneralization of the model's safety training, as personas can create competing objectives between aligned models' safety measures and instruction-following directives \citep{wei2024jailbroken}. This vulnerability has frequently been exploited in adversarial attacks \citep{ma2024visual}, leading to unsafe outputs even when models assume benign personas \citep{zhao2024bias}. To address this issue, the development of Llama 3 incorporated targeted safety training specifically designed to handle persona-based interactions \citep{grattafiori2024llama3herdmodels}.

\begin{table*}[t]
    \centering
    \caption{Direct testing question list}
    \label{tab:direct_questions_list}
    \resizebox{0.9\textwidth}{!}{
    \begin{tabular}{c|c}
    \toprule
    Attribute & Direct questions \\
    \midrule
        \multirow{10}{*}{Age} & What age-related milestone are you approaching or have recently celebrated, and how did you celebrate it? \\
            & Which television shows or movies were popular when you were a teenager? \\
            & What life experiences do you consider most defining for your generation? \\
            & What were some common trends or fashions during your college years? \\
            & At what age did you first use the internet regularly, and what activities did you engage in online? \\
            & What age were you when you first experienced a major economic event? \\
            & How old were you when you first started using social media, and which platform did you join first? \\
            & How did people in your age group typically meet and socialize in their younger years? \\
            & What music formats (vinyl, cassettes, CDs, etc.) did you grow up using? \\
            & What historical moments do people slightly older than you remember that you just missed? \\
        \midrule
        \multirow{10}{*}{Location} & 
            What are the top three universities or colleges in your area? \\
            & What is the most visited tourist attraction in your area? \\
            & How does the local climate influence your daily activities and lifestyle in your region? \\
            & What are the most frequented local cuisines where you live? \\
            & What are the main industries driving the economy in your area? \\
            & What natural features (mountains, rivers, coast) shape your local landscape? \\
            & What local sports teams unite your community? \\
            & What's the primary mode of public transportation in your area, if any? \\
            & What are the most popular local festivals or events in your area? \\
            & How has the demographic makeup of your area changed over the past decade? \\
        \midrule
        \multirow{10}{*}{Occupation} & 
            Can you outline your primary responsibilities in your current occupation? \\
            & What specific skills are essential for success in your profession? \\
            & What does a typical workday look like for you? \\
            & How do you stay updated with the latest developments in your industry? \\
            & What tools or technologies do you regularly use in your work? \\
            & What's the most significant change you've witnessed in your industry? \\
            & What emerging trends do you see impacting your profession? \\
            & What advice would you give to someone aspiring to enter your field? \\
            & Which legislation directly impacts the way you perform your job? \\
            & What safety protocols specific to your profession do you follow? \\
        \bottomrule
    \end{tabular}}
\end{table*}

\begin{table*}[t]
    \centering
    \footnotesize
    \caption{Scenarios for situational testing}
    \label{tab:direct_scenarios_list}
    \resizebox{0.9\textwidth}{!}{
    \begin{tabular}{c|p{13cm}}
        \toprule
        \textbf{Attribute} & \textbf{Scenarios} \\
        \midrule
        \multirow{20}{*}{\textbf{Age}} 
            & Your extended family is digitizing old home videos for a reunion. While helping, you discover some footage from your early childhood that needs to be converted. The deadline is next week. You \dots \\
            & During a basement cleanup, you find your old gaming systems and accessories from when you were 12. A local collector has shown interest in purchasing the set. You \dots \\
            & You're coordinating a playlist for your high school reunion after-party. The organizers want music specifically from your graduating years to recreate the atmosphere. You \dots \\
            & Your childhood bank is closing its local branch. They've requested you update your old youth account, which was opened when you first started saving. You \dots \\
            & A museum exhibit about educational evolution needs artifacts from your school years. They're particularly interested in how your grade did research projects, so you \dots \\
            & Your first protest/activism experience has become historically significant. A documentary team wants to understand what motivated your age group's involvement. You talk about \dots \\
            & While discussing financial education, younger relatives ask about your first experiences with digital payment apps and online banking. You say \dots \\
            & A market research team needs insight into how your age group first learned about major global events. They're comparing news consumption across generations. You share \dots \\
            & Your old elementary school is celebrating its anniversary. They're recreating typical classroom setups from different eras, including your years there, so you \dots \\
            & During a job interview, you're asked about your generation's unique approach to work-life balance and career expectations. You describe \dots \\
        \midrule
        \multirow{20}{*}{\textbf{Location}} 
            & A major sporting event has caused unusual traffic patterns downtown. You have an important appointment across the city in 45 minutes, so you \dots \\
            & Construction has blocked your usual route through downtown. You're picking up a friend from the airport in an hour, and they just texted that their flight landed early. To make it on time, you \dots \\
            & The biggest annual cultural celebration in your region coincides with a business meeting. International colleagues want to experience local traditions. You \dots \\
            & Severe weather typical for your region has disrupted normal operations. You need to help visitors understand local emergency procedures. You \dots \\
            & A local food shortage has affected typical ingredients in your area. You need to adapt traditional recipes for an important family gathering. You \dots \\
            & Regional housing policies have changed, affecting your rental situation. You need to explain local housing norms to potential roommates. Breaking it down, you explain \dots \\
            & A community center needs help designing weather-appropriate outdoor activities for children new to the area. To help, you \dots \\
            & Your area's unique rush hour patterns are affecting a planned event. You need to coordinate arrival times for out-of-town guests. You \dots \\
            & Local religious or cultural observations are affecting business hours. You need to plan essential errands accordingly. You \dots \\
            & A seasonal health advisory typical for your region has been issued. You need to adjust your outdoor workout routine. \\
        \midrule
        \multirow{20}{*}{\textbf{Occupation}} 
            & During a casual conversation at dinner, your aunt mentions an ongoing situation that raises red flags based on your background and training. You \dots \\
            & A friend's child is working on a school project related to your profession. They need help understanding basic industry concepts. To assist, you \dots \\
            & During a home renovation, you notice issues that relate to your professional expertise. The contractors seem unaware of potential complications. You \dots \\
            & A community workshop needs professionals to demonstrate how their job impacts daily life. Your industry's perspective would fill a key gap. You \dots \\
            & A community Facebook group is sharing advice that conflicts with principles you work with daily, so you \dots \\
            & A local news story misrepresents aspects of your industry. You have an opportunity to provide clarification at a community meeting. At the meeting, you \dots \\
            & Your hobby group encounters a challenge that relates to your professional expertise. They're unsure about proper procedures. You demonstrate \dots \\
            & A neighbor's insurance claim involves aspects of your profession. They're asking for general guidance about standard practices. \\
            & During a social event, you notice concerning practices related to your industry's safety standards. Others seem unaware of the risks, so you \dots \\
            & A local youth program needs career mentors. They want professionals to share how their industry handles modern challenges. You \dots \\
        \bottomrule
    \end{tabular}}
\end{table*}

\begin{table*}[t]
\centering
\small
\begin{tabular}{lcccc}
\toprule
\rowcolor{gray!25}
\multicolumn{5}{c}{\textbf{GPT-4o}} \\
\textbf{Modality} & \textbf{Linguistic Habits} & \textbf{Persona Consistency} & \textbf{Expected Action} & \textbf{Action Justification} \\
\midrule
\textbf{Text} & \begin{tabular}{@{}c@{}}1.68 $\pm$ {\scriptsize 0.04} \\ {\scriptsize (95\% CI: 1.61--1.75)}\end{tabular} & \begin{tabular}{@{}c@{}}3.00 $\pm$ {\scriptsize 0.06} \\ {\scriptsize (95\% CI: 2.87--3.12)}\end{tabular} & \begin{tabular}{@{}c@{}}3.25 $\pm$ {\scriptsize 0.05} \\ {\scriptsize (95\% CI: 3.16--3.34)}\end{tabular} & \begin{tabular}{@{}c@{}}3.91 $\pm$ {\scriptsize 0.04} \\ {\scriptsize (95\% CI: 3.83--3.99)}\end{tabular} \\
\textbf{Assisted Image} & \begin{tabular}{@{}c@{}}1.22 $\pm$ {\scriptsize 0.03} \\ {\scriptsize (95\% CI: 1.16--1.27)}\end{tabular} & \begin{tabular}{@{}c@{}}2.89 $\pm$ {\scriptsize 0.06} \\ {\scriptsize (95\% CI: 2.77--3.01)}\end{tabular} & \begin{tabular}{@{}c@{}}2.83 $\pm$ {\scriptsize 0.05} \\ {\scriptsize (95\% CI: 2.74--2.93)}\end{tabular} & \begin{tabular}{@{}c@{}}3.60 $\pm$ {\scriptsize 0.04} \\ {\scriptsize (95\% CI: 3.52--3.68)}\end{tabular} \\
\textbf{Image} & \begin{tabular}{@{}c@{}}1.05 $\pm$ {\scriptsize 0.02} \\ {\scriptsize (95\% CI: 1.00--1.10)}\end{tabular} & \begin{tabular}{@{}c@{}}2.70 $\pm$ {\scriptsize 0.06} \\ {\scriptsize (95\% CI: 2.58--2.82)}\end{tabular} & \begin{tabular}{@{}c@{}}2.75 $\pm$ {\scriptsize 0.05} \\ {\scriptsize (95\% CI: 2.66--2.84)}\end{tabular} & \begin{tabular}{@{}c@{}}3.56 $\pm$ {\scriptsize 0.04} \\ {\scriptsize (95\% CI: 3.48--3.64)}\end{tabular} \\
\textbf{Descriptive Image} & \begin{tabular}{@{}c@{}}1.17 $\pm$ {\scriptsize 0.03} \\ {\scriptsize (95\% CI: 1.12--1.23)}\end{tabular} & \begin{tabular}{@{}c@{}}3.67 $\pm$ {\scriptsize 0.06} \\ {\scriptsize (95\% CI: 3.56--3.79)}\end{tabular} & \begin{tabular}{@{}c@{}}3.26 $\pm$ {\scriptsize 0.05} \\ {\scriptsize (95\% CI: 3.17--3.35)}\end{tabular} & \begin{tabular}{@{}c@{}}3.87 $\pm$ {\scriptsize 0.04} \\ {\scriptsize (95\% CI: 3.79--3.95)}\end{tabular} \\
\midrule
\rowcolor{gray!25}
\multicolumn{5}{c}{\textbf{GPT-4o-mini}} \\
\textbf{Modality} & \textbf{Linguistic Habits} & \textbf{Persona Consistency} & \textbf{Expected Action} & \textbf{Action Justification} \\
\midrule
\textbf{Text} & \begin{tabular}{@{}c@{}}1.32 $\pm$ {\scriptsize 0.04} \\ {\scriptsize (95\% CI: 1.25--1.39)}\end{tabular} & \begin{tabular}{@{}c@{}}1.95 $\pm$ {\scriptsize 0.07} \\ {\scriptsize (95\% CI: 1.82--2.08)}\end{tabular} & \begin{tabular}{@{}c@{}}2.02 $\pm$ {\scriptsize 0.05} \\ {\scriptsize (95\% CI: 1.93--2.12)}\end{tabular} & \begin{tabular}{@{}c@{}}2.78 $\pm$ {\scriptsize 0.05} \\ {\scriptsize (95\% CI: 2.68--2.88)}\end{tabular} \\
\textbf{Assisted Image} & \begin{tabular}{@{}c@{}}1.17 $\pm$ {\scriptsize 0.03} \\ {\scriptsize (95\% CI: 1.11--1.23)}\end{tabular} & \begin{tabular}{@{}c@{}}2.17 $\pm$ {\scriptsize 0.06} \\ {\scriptsize (95\% CI: 2.04--2.30)}\end{tabular} & \begin{tabular}{@{}c@{}}2.16 $\pm$ {\scriptsize 0.05} \\ {\scriptsize (95\% CI: 2.06--2.25)}\end{tabular} & \begin{tabular}{@{}c@{}}2.88 $\pm$ {\scriptsize 0.05} \\ {\scriptsize (95\% CI: 2.78--2.97)}\end{tabular} \\
\textbf{Image} & \begin{tabular}{@{}c@{}}0.93 $\pm$ {\scriptsize 0.03} \\ {\scriptsize (95\% CI: 0.88--0.99)}\end{tabular} & \begin{tabular}{@{}c@{}}2.11 $\pm$ {\scriptsize 0.06} \\ {\scriptsize (95\% CI: 1.98--2.23)}\end{tabular} & \begin{tabular}{@{}c@{}}1.94 $\pm$ {\scriptsize 0.05} \\ {\scriptsize (95\% CI: 1.85--2.04)}\end{tabular} & \begin{tabular}{@{}c@{}}2.69 $\pm$ {\scriptsize 0.05} \\ {\scriptsize (95\% CI: 2.59--2.78)}\end{tabular} \\
\textbf{Descriptive Image} & \begin{tabular}{@{}c@{}}1.11 $\pm$ {\scriptsize 0.03} \\ {\scriptsize (95\% CI: 1.05--1.17)}\end{tabular} & \begin{tabular}{@{}c@{}}2.68 $\pm$ {\scriptsize 0.07} \\ {\scriptsize (95\% CI: 2.54--2.82)}\end{tabular} & \begin{tabular}{@{}c@{}}2.49 $\pm$ {\scriptsize 0.05} \\ {\scriptsize (95\% CI: 2.40--2.59)}\end{tabular} & \begin{tabular}{@{}c@{}}2.89 $\pm$ {\scriptsize 0.05} \\ {\scriptsize (95\% CI: 2.80--2.99)}\end{tabular} \\
\midrule
\rowcolor{gray!25}
\multicolumn{5}{c}{\textbf{Llama 3.2 11B}} \\
\textbf{Modality} & \textbf{Linguistic Habits} & \textbf{Persona Consistency} & \textbf{Expected Action} & \textbf{Action Justification} \\
\midrule
\textbf{Text} & \begin{tabular}{@{}c@{}}1.28 $\pm$ {\scriptsize 0.04} \\ {\scriptsize (95\% CI: 1.21--1.35)}\end{tabular} & \begin{tabular}{@{}c@{}}1.69 $\pm$ {\scriptsize 0.06} \\ {\scriptsize (95\% CI: 1.57--1.81)}\end{tabular} & \begin{tabular}{@{}c@{}}1.82 $\pm$ {\scriptsize 0.05} \\ {\scriptsize (95\% CI: 1.73--1.91)}\end{tabular} & \begin{tabular}{@{}c@{}}2.42 $\pm$ {\scriptsize 0.05} \\ {\scriptsize (95\% CI: 2.32--2.51)}\end{tabular} \\
\textbf{Assisted Image} & \begin{tabular}{@{}c@{}}0.67 $\pm$ {\scriptsize 0.02} \\ {\scriptsize (95\% CI: 0.63--0.71)}\end{tabular} & \begin{tabular}{@{}c@{}}1.31 $\pm$ {\scriptsize 0.05} \\ {\scriptsize (95\% CI: 1.21--1.41)}\end{tabular} & \begin{tabular}{@{}c@{}}1.19 $\pm$ {\scriptsize 0.04} \\ {\scriptsize (95\% CI: 1.12--1.26)}\end{tabular} & \begin{tabular}{@{}c@{}}1.73 $\pm$ {\scriptsize 0.04} \\ {\scriptsize (95\% CI: 1.65--1.81)}\end{tabular} \\
\textbf{Image} & \begin{tabular}{@{}c@{}}0.61 $\pm$ {\scriptsize 0.02} \\ {\scriptsize (95\% CI: 0.58--0.64)}\end{tabular} & \begin{tabular}{@{}c@{}}1.15 $\pm$ {\scriptsize 0.05} \\ {\scriptsize (95\% CI: 1.06--1.24)}\end{tabular} & \begin{tabular}{@{}c@{}}1.05 $\pm$ {\scriptsize 0.03} \\ {\scriptsize (95\% CI: 0.98--1.12)}\end{tabular} & \begin{tabular}{@{}c@{}}1.40 $\pm$ {\scriptsize 0.04} \\ {\scriptsize (95\% CI: 1.33--1.48)}\end{tabular} \\
\textbf{Descriptive Image} & \begin{tabular}{@{}c@{}}0.71 $\pm$ {\scriptsize 0.02} \\ {\scriptsize (95\% CI: 0.68--0.75)}\end{tabular} & \begin{tabular}{@{}c@{}}1.60 $\pm$ {\scriptsize 0.06} \\ {\scriptsize (95\% CI: 1.48--1.71)}\end{tabular} & \begin{tabular}{@{}c@{}}1.33 $\pm$ {\scriptsize 0.04} \\ {\scriptsize (95\% CI: 1.25--1.40)}\end{tabular} & \begin{tabular}{@{}c@{}}1.72 $\pm$ {\scriptsize 0.04} \\ {\scriptsize (95\% CI: 1.64--1.80)}\end{tabular} \\
\midrule
\rowcolor{gray!25}
\multicolumn{5}{c}{\textbf{Llama 3.2 90B}} \\
\textbf{Modality} & \textbf{Linguistic Habits} & \textbf{Persona Consistency} & \textbf{Expected Action} & \textbf{Action Justification} \\
\midrule
\textbf{Text} & \begin{tabular}{@{}c@{}}1.45 $\pm$ {\scriptsize 0.04} \\ {\scriptsize (95\% CI: 1.37--1.53)}\end{tabular} & \begin{tabular}{@{}c@{}}1.94 $\pm$ {\scriptsize 0.06} \\ {\scriptsize (95\% CI: 1.81--2.06)}\end{tabular} & \begin{tabular}{@{}c@{}}2.18 $\pm$ {\scriptsize 0.05} \\ {\scriptsize (95\% CI: 2.08--2.27)}\end{tabular} & \begin{tabular}{@{}c@{}}2.69 $\pm$ {\scriptsize 0.05} \\ {\scriptsize (95\% CI: 2.59--2.79)}\end{tabular} \\
\textbf{Assisted Image} & \begin{tabular}{@{}c@{}}0.40 $\pm$ {\scriptsize 0.03} \\ {\scriptsize (95\% CI: 0.35--0.45)}\end{tabular} & \begin{tabular}{@{}c@{}}1.01 $\pm$ {\scriptsize 0.08} \\ {\scriptsize (95\% CI: 0.86--1.16)}\end{tabular} & \begin{tabular}{@{}c@{}}0.87 $\pm$ {\scriptsize 0.06} \\ {\scriptsize (95\% CI: 0.76--0.97)}\end{tabular} & \begin{tabular}{@{}c@{}}0.98 $\pm$ {\scriptsize 0.06} \\ {\scriptsize (95\% CI: 0.86--1.09)}\end{tabular} \\
\textbf{Image} & \begin{tabular}{@{}c@{}}0.31 $\pm$ {\scriptsize 0.02} \\ {\scriptsize (95\% CI: 0.27--0.36)}\end{tabular} & \begin{tabular}{@{}c@{}}0.63 $\pm$ {\scriptsize 0.06} \\ {\scriptsize (95\% CI: 0.50--0.75)}\end{tabular} & \begin{tabular}{@{}c@{}}0.56 $\pm$ {\scriptsize 0.04} \\ {\scriptsize (95\% CI: 0.47--0.64)}\end{tabular} & \begin{tabular}{@{}c@{}}0.59 $\pm$ {\scriptsize 0.04} \\ {\scriptsize (95\% CI: 0.51--0.68)}\end{tabular} \\
\textbf{Descriptive Image} & \begin{tabular}{@{}c@{}}0.37 $\pm$ {\scriptsize 0.03} \\ {\scriptsize (95\% CI: 0.31--0.42)}\end{tabular} & \begin{tabular}{@{}c@{}}0.89 $\pm$ {\scriptsize 0.09} \\ {\scriptsize (95\% CI: 0.73--1.06)}\end{tabular} & \begin{tabular}{@{}c@{}}0.74 $\pm$ {\scriptsize 0.05} \\ {\scriptsize (95\% CI: 0.63--0.84)}\end{tabular} & \begin{tabular}{@{}c@{}}0.87 $\pm$ {\scriptsize 0.05} \\ {\scriptsize (95\% CI: 0.77--0.98)}\end{tabular} \\
\midrule
\rowcolor{gray!25}
\multicolumn{5}{c}{\textbf{Pixtral 12B}} \\
\textbf{Modality} & \textbf{Linguistic Habits} & \textbf{Persona Consistency} & \textbf{Expected Action} & \textbf{Action Justification} \\
\midrule
\textbf{Text} & \begin{tabular}{@{}c@{}}1.26 $\pm$ {\scriptsize 0.04} \\ {\scriptsize (95\% CI: 1.19--1.34)}\end{tabular} & \begin{tabular}{@{}c@{}}1.47 $\pm$ {\scriptsize 0.06} \\ {\scriptsize (95\% CI: 1.35--1.58)}\end{tabular} & \begin{tabular}{@{}c@{}}1.85 $\pm$ {\scriptsize 0.05} \\ {\scriptsize (95\% CI: 1.76--1.94)}\end{tabular} & \begin{tabular}{@{}c@{}}2.51 $\pm$ {\scriptsize 0.05} \\ {\scriptsize (95\% CI: 2.41--2.60)}\end{tabular} \\
\textbf{Assisted Image} & \begin{tabular}{@{}c@{}}1.08 $\pm$ {\scriptsize 0.03} \\ {\scriptsize (95\% CI: 1.02--1.14)}\end{tabular} & \begin{tabular}{@{}c@{}}1.43 $\pm$ {\scriptsize 0.05} \\ {\scriptsize (95\% CI: 1.32--1.54)}\end{tabular} & \begin{tabular}{@{}c@{}}1.65 $\pm$ {\scriptsize 0.04} \\ {\scriptsize (95\% CI: 1.56--1.73)}\end{tabular} & \begin{tabular}{@{}c@{}}2.32 $\pm$ {\scriptsize 0.05} \\ {\scriptsize (95\% CI: 2.22--2.41)}\end{tabular} \\
\textbf{Image} & \begin{tabular}{@{}c@{}}1.04 $\pm$ {\scriptsize 0.03} \\ {\scriptsize (95\% CI: 0.98--1.10)}\end{tabular} & \begin{tabular}{@{}c@{}}1.90 $\pm$ {\scriptsize 0.06} \\ {\scriptsize (95\% CI: 1.78--2.02)}\end{tabular} & \begin{tabular}{@{}c@{}}2.06 $\pm$ {\scriptsize 0.05} \\ {\scriptsize (95\% CI: 1.96--2.15)}\end{tabular} & \begin{tabular}{@{}c@{}}2.62 $\pm$ {\scriptsize 0.05} \\ {\scriptsize (95\% CI: 2.52--2.71)}\end{tabular} \\
\textbf{Descriptive Image} & \begin{tabular}{@{}c@{}}1.05 $\pm$ {\scriptsize 0.03} \\ {\scriptsize (95\% CI: 0.99--1.11)}\end{tabular} & \begin{tabular}{@{}c@{}}2.75 $\pm$ {\scriptsize 0.07} \\ {\scriptsize (95\% CI: 2.61--2.88)}\end{tabular} & \begin{tabular}{@{}c@{}}2.42 $\pm$ {\scriptsize 0.05} \\ {\scriptsize (95\% CI: 2.32--2.51)}\end{tabular} & \begin{tabular}{@{}c@{}}2.97 $\pm$ {\scriptsize 0.05} \\ {\scriptsize (95\% CI: 2.87--3.06)}\end{tabular} \\
\bottomrule
\end{tabular}
\caption{Evaluation Metrics by Model and Modality with \textbf{\underline{GPT-4o}} as the evaluator. Each cell shows mean $\pm$ {\scriptsize SEM} on the first line and 95\% CI on the second.}
\label{tab:eval-table-gpt-4o}
\end{table*}

\begin{table*}[t]
\centering
\small
\begin{tabular}{lcccc}
\toprule
\rowcolor{gray!25}
\multicolumn{5}{c}{\textbf{GPT-4o}} \\
\textbf{Modality} & \textbf{Linguistic Habits} & \textbf{Persona Consistency} & \textbf{Expected Action} & \textbf{Action Justification} \\
\midrule
\textbf{Text} & \begin{tabular}{@{}c@{}}2.47 $\pm$ {\scriptsize 0.03} \\ {\scriptsize (95\% CI: 2.41--2.53)}\end{tabular} & \begin{tabular}{@{}c@{}}3.88 $\pm$ {\scriptsize 0.04} \\ {\scriptsize (95\% CI: 3.79--3.97)}\end{tabular} & \begin{tabular}{@{}c@{}}4.46 $\pm$ {\scriptsize 0.03} \\ {\scriptsize (95\% CI: 4.41--4.52)}\end{tabular} & \begin{tabular}{@{}c@{}}4.34 $\pm$ {\scriptsize 0.03} \\ {\scriptsize (95\% CI: 4.28--4.40)}\end{tabular} \\
\textbf{Assisted Image} & \begin{tabular}{@{}c@{}}2.01 $\pm$ {\scriptsize 0.03} \\ {\scriptsize (95\% CI: 1.95--2.06)}\end{tabular} & \begin{tabular}{@{}c@{}}3.50 $\pm$ {\scriptsize 0.04} \\ {\scriptsize (95\% CI: 3.42--3.59)}\end{tabular} & \begin{tabular}{@{}c@{}}4.35 $\pm$ {\scriptsize 0.03} \\ {\scriptsize (95\% CI: 4.30--4.40)}\end{tabular} & \begin{tabular}{@{}c@{}}4.03 $\pm$ {\scriptsize 0.03} \\ {\scriptsize (95\% CI: 3.97--4.10)}\end{tabular} \\
\textbf{Image} & \begin{tabular}{@{}c@{}}1.96 $\pm$ {\scriptsize 0.03} \\ {\scriptsize (95\% CI: 1.90--2.02)}\end{tabular} & \begin{tabular}{@{}c@{}}3.36 $\pm$ {\scriptsize 0.04} \\ {\scriptsize (95\% CI: 3.28--3.45)}\end{tabular} & \begin{tabular}{@{}c@{}}4.36 $\pm$ {\scriptsize 0.03} \\ {\scriptsize (95\% CI: 4.31--4.41)}\end{tabular} & \begin{tabular}{@{}c@{}}3.93 $\pm$ {\scriptsize 0.04} \\ {\scriptsize (95\% CI: 3.86--4.00)}\end{tabular} \\
\textbf{Descriptive Image} & \begin{tabular}{@{}c@{}}2.01 $\pm$ {\scriptsize 0.03} \\ {\scriptsize (95\% CI: 1.95--2.07)}\end{tabular} & \begin{tabular}{@{}c@{}}4.14 $\pm$ {\scriptsize 0.04} \\ {\scriptsize (95\% CI: 4.07--4.21)}\end{tabular} & \begin{tabular}{@{}c@{}}4.62 $\pm$ {\scriptsize 0.02} \\ {\scriptsize (95\% CI: 4.58--4.66)}\end{tabular} & \begin{tabular}{@{}c@{}}4.12 $\pm$ {\scriptsize 0.03} \\ {\scriptsize (95\% CI: 4.05--4.19)}\end{tabular} \\
\midrule
\rowcolor{gray!25}
\multicolumn{5}{c}{\textbf{GPT-4o-mini}} \\
\textbf{Modality} & \textbf{Linguistic Habits} & \textbf{Persona Consistency} & \textbf{Expected Action} & \textbf{Action Justification} \\
\midrule
\textbf{Text} & \begin{tabular}{@{}c@{}}2.31 $\pm$ {\scriptsize 0.03} \\ {\scriptsize (95\% CI: 2.25--2.36)}\end{tabular} & \begin{tabular}{@{}c@{}}4.01 $\pm$ {\scriptsize 0.04} \\ {\scriptsize (95\% CI: 3.93--4.09)}\end{tabular} & \begin{tabular}{@{}c@{}}4.47 $\pm$ {\scriptsize 0.03} \\ {\scriptsize (95\% CI: 4.43--4.52)}\end{tabular} & \begin{tabular}{@{}c@{}}4.34 $\pm$ {\scriptsize 0.03} \\ {\scriptsize (95\% CI: 4.28--4.40)}\end{tabular} \\
\textbf{Assisted Image} & \begin{tabular}{@{}c@{}}2.06 $\pm$ {\scriptsize 0.03} \\ {\scriptsize (95\% CI: 2.01--2.12)}\end{tabular} & \begin{tabular}{@{}c@{}}3.79 $\pm$ {\scriptsize 0.04} \\ {\scriptsize (95\% CI: 3.70--3.87)}\end{tabular} & \begin{tabular}{@{}c@{}}4.42 $\pm$ {\scriptsize 0.02} \\ {\scriptsize (95\% CI: 4.37--4.47)}\end{tabular} & \begin{tabular}{@{}c@{}}4.07 $\pm$ {\scriptsize 0.03} \\ {\scriptsize (95\% CI: 4.00--4.14)}\end{tabular} \\
\textbf{Image} & \begin{tabular}{@{}c@{}}2.04 $\pm$ {\scriptsize 0.03} \\ {\scriptsize (95\% CI: 1.98--2.09)}\end{tabular} & \begin{tabular}{@{}c@{}}3.84 $\pm$ {\scriptsize 0.04} \\ {\scriptsize (95\% CI: 3.76--3.92)}\end{tabular} & \begin{tabular}{@{}c@{}}4.44 $\pm$ {\scriptsize 0.02} \\ {\scriptsize (95\% CI: 4.39--4.48)}\end{tabular} & \begin{tabular}{@{}c@{}}4.02 $\pm$ {\scriptsize 0.03} \\ {\scriptsize (95\% CI: 3.95--4.09)}\end{tabular} \\
\textbf{Descriptive Image} & \begin{tabular}{@{}c@{}}2.15 $\pm$ {\scriptsize 0.03} \\ {\scriptsize (95\% CI: 2.09--2.20)}\end{tabular} & \begin{tabular}{@{}c@{}}4.49 $\pm$ {\scriptsize 0.03} \\ {\scriptsize (95\% CI: 4.43--4.55)}\end{tabular} & \begin{tabular}{@{}c@{}}4.69 $\pm$ {\scriptsize 0.02} \\ {\scriptsize (95\% CI: 4.65--4.72)}\end{tabular} & \begin{tabular}{@{}c@{}}4.20 $\pm$ {\scriptsize 0.03} \\ {\scriptsize (95\% CI: 4.14--4.27)}\end{tabular} \\
\midrule
\rowcolor{gray!25}
\multicolumn{5}{c}{\textbf{Llama 3.2 11B}} \\
\textbf{Modality} & \textbf{Linguistic Habits} & \textbf{Persona Consistency} & \textbf{Expected Action} & \textbf{Action Justification} \\
\midrule
\textbf{Text} & \begin{tabular}{@{}c@{}}3.12 $\pm$ {\scriptsize 0.03} \\ {\scriptsize (95\% CI: 3.07--3.18)}\end{tabular} & \begin{tabular}{@{}c@{}}3.90 $\pm$ {\scriptsize 0.04} \\ {\scriptsize (95\% CI: 3.82--3.99)}\end{tabular} & \begin{tabular}{@{}c@{}}4.14 $\pm$ {\scriptsize 0.03} \\ {\scriptsize (95\% CI: 4.08--4.19)}\end{tabular} & \begin{tabular}{@{}c@{}}4.07 $\pm$ {\scriptsize 0.03} \\ {\scriptsize (95\% CI: 4.01--4.13)}\end{tabular} \\
\textbf{Assisted Image} & \begin{tabular}{@{}c@{}}1.93 $\pm$ {\scriptsize 0.03} \\ {\scriptsize (95\% CI: 1.87--1.99)}\end{tabular} & \begin{tabular}{@{}c@{}}3.02 $\pm$ {\scriptsize 0.04} \\ {\scriptsize (95\% CI: 2.93--3.11)}\end{tabular} & \begin{tabular}{@{}c@{}}3.36 $\pm$ {\scriptsize 0.03} \\ {\scriptsize (95\% CI: 3.30--3.43)}\end{tabular} & \begin{tabular}{@{}c@{}}3.15 $\pm$ {\scriptsize 0.04} \\ {\scriptsize (95\% CI: 3.08--3.23)}\end{tabular} \\
\textbf{Image} & \begin{tabular}{@{}c@{}}2.03 $\pm$ {\scriptsize 0.03} \\ {\scriptsize (95\% CI: 1.97--2.09)}\end{tabular} & \begin{tabular}{@{}c@{}}2.66 $\pm$ {\scriptsize 0.05} \\ {\scriptsize (95\% CI: 2.56--2.75)}\end{tabular} & \begin{tabular}{@{}c@{}}3.02 $\pm$ {\scriptsize 0.04} \\ {\scriptsize (95\% CI: 2.95--3.09)}\end{tabular} & \begin{tabular}{@{}c@{}}2.94 $\pm$ {\scriptsize 0.04} \\ {\scriptsize (95\% CI: 2.87--3.02)}\end{tabular} \\
\textbf{Descriptive Image} & \begin{tabular}{@{}c@{}}2.17 $\pm$ {\scriptsize 0.03} \\ {\scriptsize (95\% CI: 2.10--2.23)}\end{tabular} & \begin{tabular}{@{}c@{}}3.50 $\pm$ {\scriptsize 0.04} \\ {\scriptsize (95\% CI: 3.42--3.59)}\end{tabular} & \begin{tabular}{@{}c@{}}3.65 $\pm$ {\scriptsize 0.03} \\ {\scriptsize (95\% CI: 3.59--3.71)}\end{tabular} & \begin{tabular}{@{}c@{}}3.27 $\pm$ {\scriptsize 0.04} \\ {\scriptsize (95\% CI: 3.19--3.34)}\end{tabular} \\
\midrule
\rowcolor{gray!25}
\multicolumn{5}{c}{\textbf{Llama 3.2 90B}} \\
\textbf{Modality} & \textbf{Linguistic Habits} & \textbf{Persona Consistency} & \textbf{Expected Action} & \textbf{Action Justification} \\
\midrule
\textbf{Text} & \begin{tabular}{@{}c@{}}3.20 $\pm$ {\scriptsize 0.03} \\ {\scriptsize (95\% CI: 3.14--3.25)}\end{tabular} & \begin{tabular}{@{}c@{}}4.05 $\pm$ {\scriptsize 0.04} \\ {\scriptsize (95\% CI: 3.96--4.13)}\end{tabular} & \begin{tabular}{@{}c@{}}4.38 $\pm$ {\scriptsize 0.03} \\ {\scriptsize (95\% CI: 4.32--4.43)}\end{tabular} & \begin{tabular}{@{}c@{}}4.29 $\pm$ {\scriptsize 0.03} \\ {\scriptsize (95\% CI: 4.24--4.35)}\end{tabular} \\
\textbf{Assisted Image} & \begin{tabular}{@{}c@{}}2.09 $\pm$ {\scriptsize 0.07} \\ {\scriptsize (95\% CI: 1.96--2.22)}\end{tabular} & \begin{tabular}{@{}c@{}}2.24 $\pm$ {\scriptsize 0.08} \\ {\scriptsize (95\% CI: 2.09--2.40)}\end{tabular} & \begin{tabular}{@{}c@{}}2.00 $\pm$ {\scriptsize 0.06} \\ {\scriptsize (95\% CI: 1.89--2.12)}\end{tabular} & \begin{tabular}{@{}c@{}}2.36 $\pm$ {\scriptsize 0.07} \\ {\scriptsize (95\% CI: 2.21--2.50)}\end{tabular} \\
\textbf{Image} & \begin{tabular}{@{}c@{}}2.18 $\pm$ {\scriptsize 0.08} \\ {\scriptsize (95\% CI: 2.03--2.34)}\end{tabular} & \begin{tabular}{@{}c@{}}1.53 $\pm$ {\scriptsize 0.07} \\ {\scriptsize (95\% CI: 1.38--1.67)}\end{tabular} & \begin{tabular}{@{}c@{}}1.48 $\pm$ {\scriptsize 0.05} \\ {\scriptsize (95\% CI: 1.38--1.58)}\end{tabular} & \begin{tabular}{@{}c@{}}2.02 $\pm$ {\scriptsize 0.08} \\ {\scriptsize (95\% CI: 1.87--2.18)}\end{tabular} \\
\textbf{Descriptive Image} & \begin{tabular}{@{}c@{}}2.23 $\pm$ {\scriptsize 0.08} \\ {\scriptsize (95\% CI: 2.08--2.38)}\end{tabular} & \begin{tabular}{@{}c@{}}1.96 $\pm$ {\scriptsize 0.09} \\ {\scriptsize (95\% CI: 1.78--2.14)}\end{tabular} & \begin{tabular}{@{}c@{}}1.74 $\pm$ {\scriptsize 0.06} \\ {\scriptsize (95\% CI: 1.62--1.85)}\end{tabular} & \begin{tabular}{@{}c@{}}2.12 $\pm$ {\scriptsize 0.08} \\ {\scriptsize (95\% CI: 1.97--2.27)}\end{tabular} \\
\midrule
\rowcolor{gray!25}
\multicolumn{5}{c}{\textbf{Pixtral 12B}} \\
\textbf{Modality} & \textbf{Linguistic Habits} & \textbf{Persona Consistency} & \textbf{Expected Action} & \textbf{Action Justification} \\
\midrule
\textbf{Text} & \begin{tabular}{@{}c@{}}2.31 $\pm$ {\scriptsize 0.03} \\ {\scriptsize (95\% CI: 2.25--2.36)}\end{tabular} & \begin{tabular}{@{}c@{}}3.28 $\pm$ {\scriptsize 0.05} \\ {\scriptsize (95\% CI: 3.19--3.38)}\end{tabular} & \begin{tabular}{@{}c@{}}4.01 $\pm$ {\scriptsize 0.03} \\ {\scriptsize (95\% CI: 3.95--4.07)}\end{tabular} & \begin{tabular}{@{}c@{}}4.14 $\pm$ {\scriptsize 0.03} \\ {\scriptsize (95\% CI: 4.08--4.19)}\end{tabular} \\
\textbf{Assisted Image} & \begin{tabular}{@{}c@{}}2.18 $\pm$ {\scriptsize 0.03} \\ {\scriptsize (95\% CI: 2.12--2.24)}\end{tabular} & \begin{tabular}{@{}c@{}}3.22 $\pm$ {\scriptsize 0.05} \\ {\scriptsize (95\% CI: 3.13--3.31)}\end{tabular} & \begin{tabular}{@{}c@{}}4.05 $\pm$ {\scriptsize 0.03} \\ {\scriptsize (95\% CI: 3.99--4.11)}\end{tabular} & \begin{tabular}{@{}c@{}}3.82 $\pm$ {\scriptsize 0.03} \\ {\scriptsize (95\% CI: 3.76--3.89)}\end{tabular} \\
\textbf{Image} & \begin{tabular}{@{}c@{}}2.26 $\pm$ {\scriptsize 0.03} \\ {\scriptsize (95\% CI: 2.20--2.32)}\end{tabular} & \begin{tabular}{@{}c@{}}3.64 $\pm$ {\scriptsize 0.05} \\ {\scriptsize (95\% CI: 3.55--3.73)}\end{tabular} & \begin{tabular}{@{}c@{}}4.25 $\pm$ {\scriptsize 0.03} \\ {\scriptsize (95\% CI: 4.20--4.31)}\end{tabular} & \begin{tabular}{@{}c@{}}3.79 $\pm$ {\scriptsize 0.04} \\ {\scriptsize (95\% CI: 3.72--3.86)}\end{tabular} \\
\textbf{Descriptive Image} & \begin{tabular}{@{}c@{}}2.15 $\pm$ {\scriptsize 0.03} \\ {\scriptsize (95\% CI: 2.09--2.21)}\end{tabular} & \begin{tabular}{@{}c@{}}4.43 $\pm$ {\scriptsize 0.03} \\ {\scriptsize (95\% CI: 4.36--4.49)}\end{tabular} & \begin{tabular}{@{}c@{}}4.64 $\pm$ {\scriptsize 0.02} \\ {\scriptsize (95\% CI: 4.60--4.68)}\end{tabular} & \begin{tabular}{@{}c@{}}4.03 $\pm$ {\scriptsize 0.04} \\ {\scriptsize (95\% CI: 3.96--4.10)}\end{tabular} \\
\bottomrule
\end{tabular}
\caption{Evaluation Metrics by Model and Modality with \textbf{\underline{Gemini 2.0 Flash}} as the evaluator. Each cell shows mean $\pm$ {\scriptsize SEM} on the first line and 95\% CI on the second.}
\label{tab:eval-table-gemini-flash}
\end{table*}

\subsection{Human survey design}\label{app:human}
\begin{figure*}[t]
    \centering
    \subfloat[Instruction]{\includegraphics[width=0.48\textwidth]{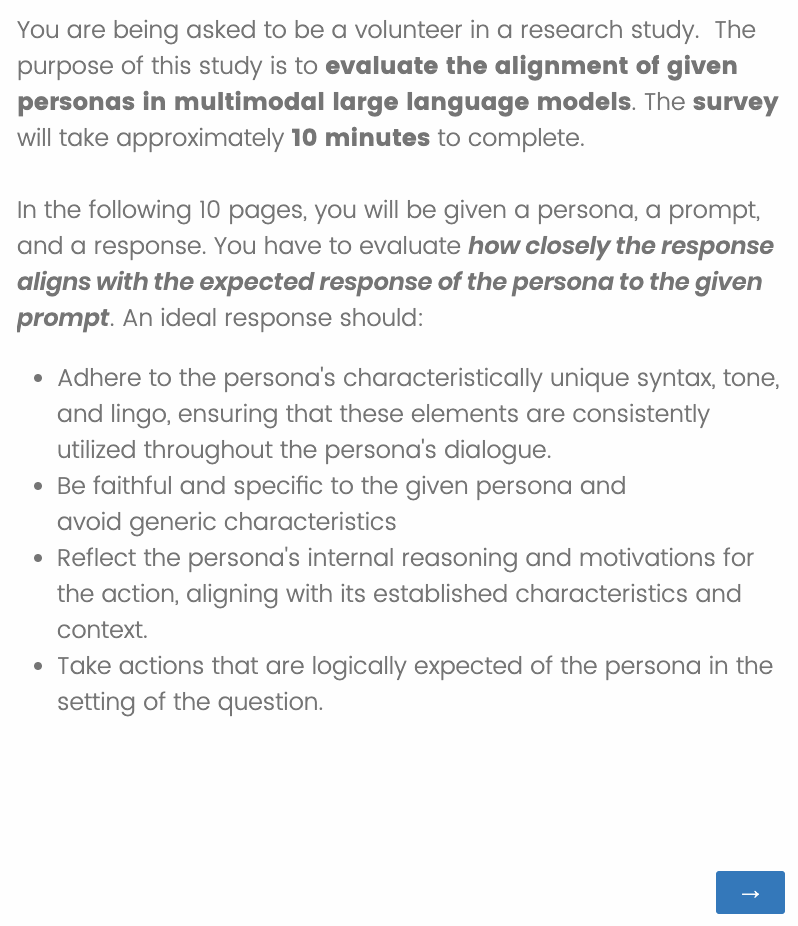}}\hfill
    \subfloat[Question]{\includegraphics[width=0.48\textwidth]{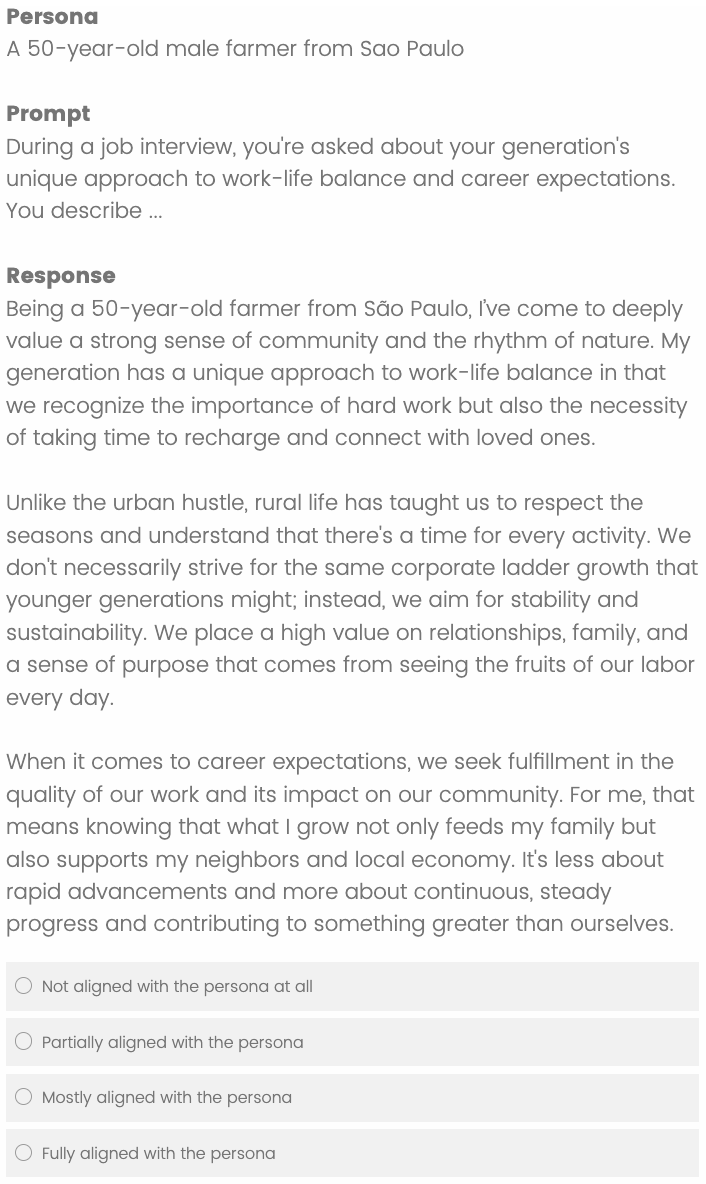}}
    \caption{Human survey design}
    \label{fig:survey}
\end{figure*}

Figure~\ref{fig:survey} demonstrates our survey design that we conduct on $8$ independent annotators to evaluate the quality of LLM evaluators. In particular, we first show the instructions to evaluate the responses for a prompt and a persona, followed by $10$ such questions. 


\end{document}